\documentclass[runningheads]{llncs}

\usepackage{eccv}

\usepackage{eccvabbrv}

\usepackage{graphicx}
\usepackage{booktabs}
\usepackage{subcaption}
\usepackage{multirow}
\usepackage{caption}
\usepackage{wrapstuff}

\usepackage[accsupp]{axessibility}  %

\usepackage{amsmath}
\usepackage{amssymb}
\usepackage{mathtools}

\usepackage[pagebackref,breaklinks,colorlinks,citecolor=eccvblue]{hyperref}

\usepackage{booktabs}
\usepackage{multirow}
\usepackage{colortbl}
\usepackage{xcolor}

\usepackage{orcidlink}

\begin{document}
\title{HSG: Hyperbolic Scene Graph} 

\titlerunning{HSG: Hyperbolic Scene Graph}

\author{
  \textbf{Liyang Wang$^{*}$~~
  Zeyu Zhang$^{*\dag}$~~
  Hao Tang$^{\ddag}$}\\
  \vspace{.5em}
  School of Computer Science, Peking University\\
  \vspace{.5em}
  \small$^{*}$Equal contribution. $^{\dag}$Project lead. $^{\ddag}$Corresponding author: bjdxtanghao@gmail.com.
}

\authorrunning{Liyang Wang, Zeyu Zhang, and Hao Tang}

\institute{}

\maketitle

\begin{abstract}
  Scene graph representations enable structured visual understanding by modeling objects and their relationships, and have been widely used for multiview and 3D scene reasoning. Existing methods such as MSG learn scene graph embeddings in Euclidean space using contrastive learning and attention-based association. However, Euclidean geometry does not explicitly capture hierarchical entailment relationships between places and objects, limiting the structural consistency of learned representations. To address this, we propose Hyperbolic Scene Graph (HSG), which learns scene graph embeddings in hyperbolic space where hierarchical relationships are naturally encoded through geometric distance. Our results show that HSG improves hierarchical structure quality while maintaining strong retrieval performance. The largest gains are observed in graph-level metrics: HSG achieves a PP IoU of \textbf{33.17} and the highest Graph IoU of \textbf{33.51}, outperforming the best AoMSG variant (25.37) by \textbf{8.14}, highlighting the effectiveness of hyperbolic representation learning for scene graph modeling.
  Code:~\url{https://github.com/AIGeeksGroup/HSG}.

\end{abstract}

\section{Introduction}
\label{sec:intro}
Understanding and representing complex 3D environments is a fundamental problem in computer vision, with broad applications in robotics, autonomous navigation, and embodied AI. Scene graphs have emerged as an effective structured representation to model objects and their semantic and spatial relationships within a scene~\cite{johnson2015image, yang2018graph}. Recent advances extend scene graph representations to large-scale and long-term environments by leveraging visual place recognition and object association, enabling consistent scene understanding across viewpoint and temporal variations. In particular, MSG-based methods~\cite{zhang2024multiview} construct multi-layer scene graphs by jointly modeling place and object nodes using learned visual embeddings. These approaches typically rely on deep visual encoders such as ConvNeXt~\cite{liu2022convnet} and DINOv2~\cite{oquab2023dinov2}, combined with contrastive learning objectives to align semantically related entities in embedding space. Despite their effectiveness, most existing methods learn embeddings in Euclidean space, which may not naturally capture the hierarchical structure inherent in real-world environments.

However, real-world scenes exhibit strong hierarchical relationships\cite{vendrov2015order}, where places semantically entail the presence of objects, and objects further exhibit a hierarchical semantic structure. Euclidean embeddings are limited in their ability to represent such hierarchical and entailment relationships efficiently, often requiring higher dimensions or resulting in suboptimal structural organization~\cite{nickel2017poincare, gu2018learning}. As a result, existing methods may achieve high place recognition accuracy while failing to preserve meaningful structural relationships in the learned embedding space, leading to suboptimal scene graph quality. 

\begin{figure}[t]
    \centering
    \includegraphics[width=\linewidth]{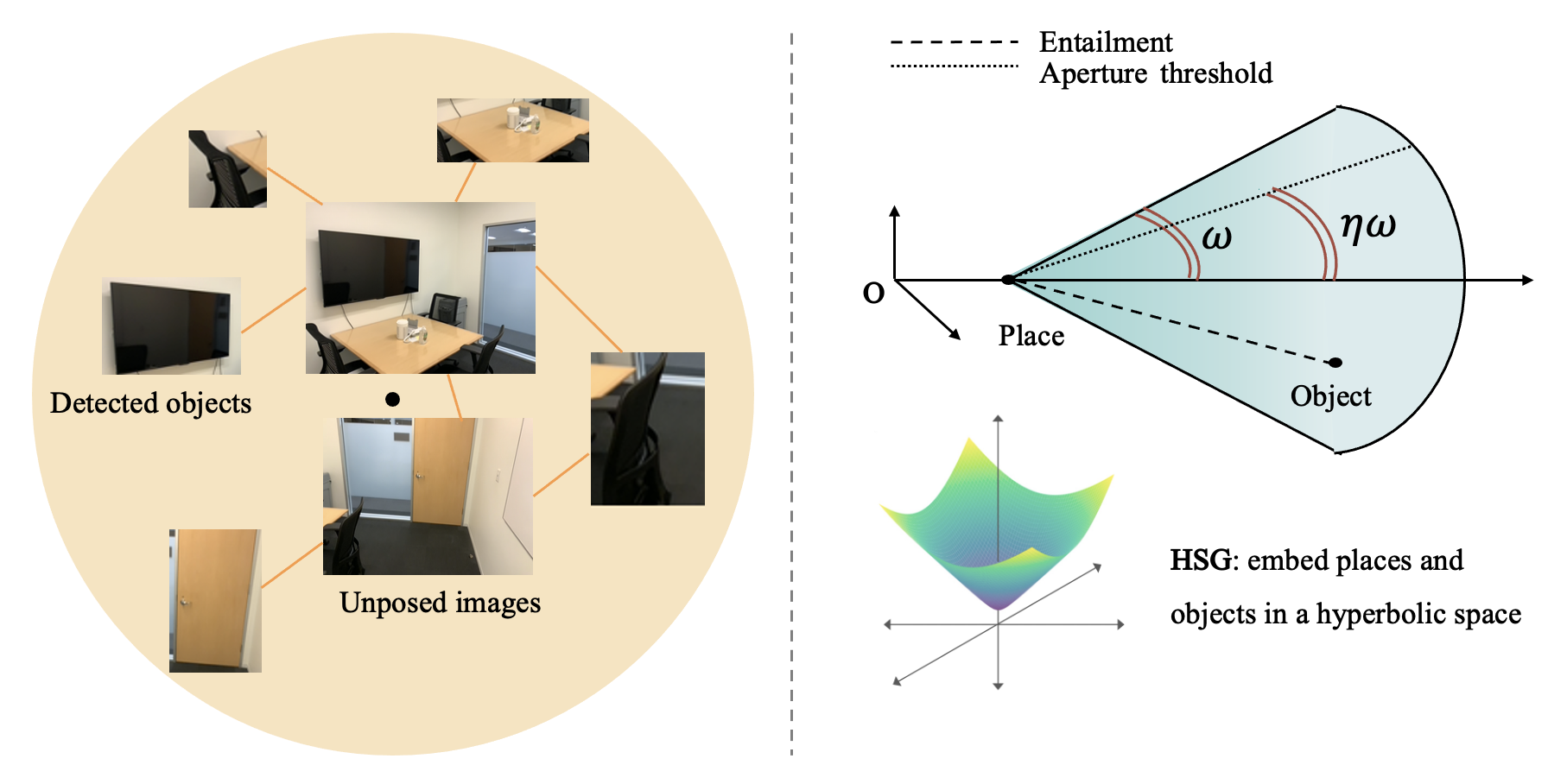}
    \caption{\textbf{Hyperbolic place-object representations.} \textbf{Left:} A place-centric visual-semantic hierarchy where a central scene representation (black dot) entails multiple object-level observations (surrounding images), illustrating the hierarchical relationship between place and object concepts. 
\textbf{Right:} Hyperbolic entailment cone formulation. The place embedding lies closer to the origin and defines an entailment cone within which valid object embeddings are constrained. The cone aperture $\omega$ controls the angular spread of entailment, while the aperture threshold $\eta$ scales the aperture $\omega$ to increase or decrease the width of the entailment cone, thereby regulating the strength of hierarchical constraints in hyperbolic space.
}
    \label{fig:visualisation}
    \vspace{-0.4cm}
\end{figure}

To address these challenges, we are motivated to explore hyperbolic representation learning, which provides an inherently hierarchical geometry with exponentially increasing capacity as a function of distance from the origin. This property makes hyperbolic space particularly suitable for modeling entailment and hierarchical relationships in scene understanding tasks, as demonstrated by recent works\cite{sala2018representation, yang2023hyperbolic, leng2024hypersdfusion, chami2019hyperbolic, zhang2026geoworld}. 

In this paper, we propose Hyperbolic Scene Graph (HSG), a novel framework that learns scene graph representations in hyperbolic space. Our approach embeds places and objects into a Lorentz hyperbolic manifold, where the origin naturally represents the most abstract concept, and more specific entities lie farther away. We further introduce an entailment loss that explicitly enforces hierarchical constraints between place and object embeddings, encouraging structurally meaningful organization in the learned space. After training, embeddings are mapped back to the tangent space via the logarithmic map, ensuring geometric consistency while enabling seamless integration with the MSG graph construction pipeline. Extensive experiments demonstrate that HSG significantly improves structural metrics such as PP IoU and Graph IoU while maintaining competitive Recall@1 compared to strong Euclidean baselines including SepMSG and AoMSG. Qualitative analysis further confirms that HSG produces embedding distributions with clear hierarchical organization aligned with semantic entailment.

In summary, the contributions of our paper can be summarized as follows:
\begin{itemize}
    \item We propose Hyperbolic Scene Graph (HSG), a novel framework that leverages hyperbolic geometry to learn hierarchical scene representations for scene graph construction.
    
    \item We introduce an entailment loss in hyperbolic space that explicitly models place-object hierarchical relationships, resulting in improved structural consistency and graph quality.
    
    \item Extensive quantitative and qualitative evaluations demonstrate that HSG produces more structured embeddings and outperforms Euclidean baselines in scene graph construction. The largest gains are in graph-level metrics: HSG achieves a PP IoU of \textbf{33.17} and a Graph IoU of \textbf{33.51}, surpassing the best AoMSG variant (25.37) by \textbf{8.14}.
\end{itemize}

\section{Related Work}
\begin{figure}[t]
    \centering
    \includegraphics[width=\linewidth]{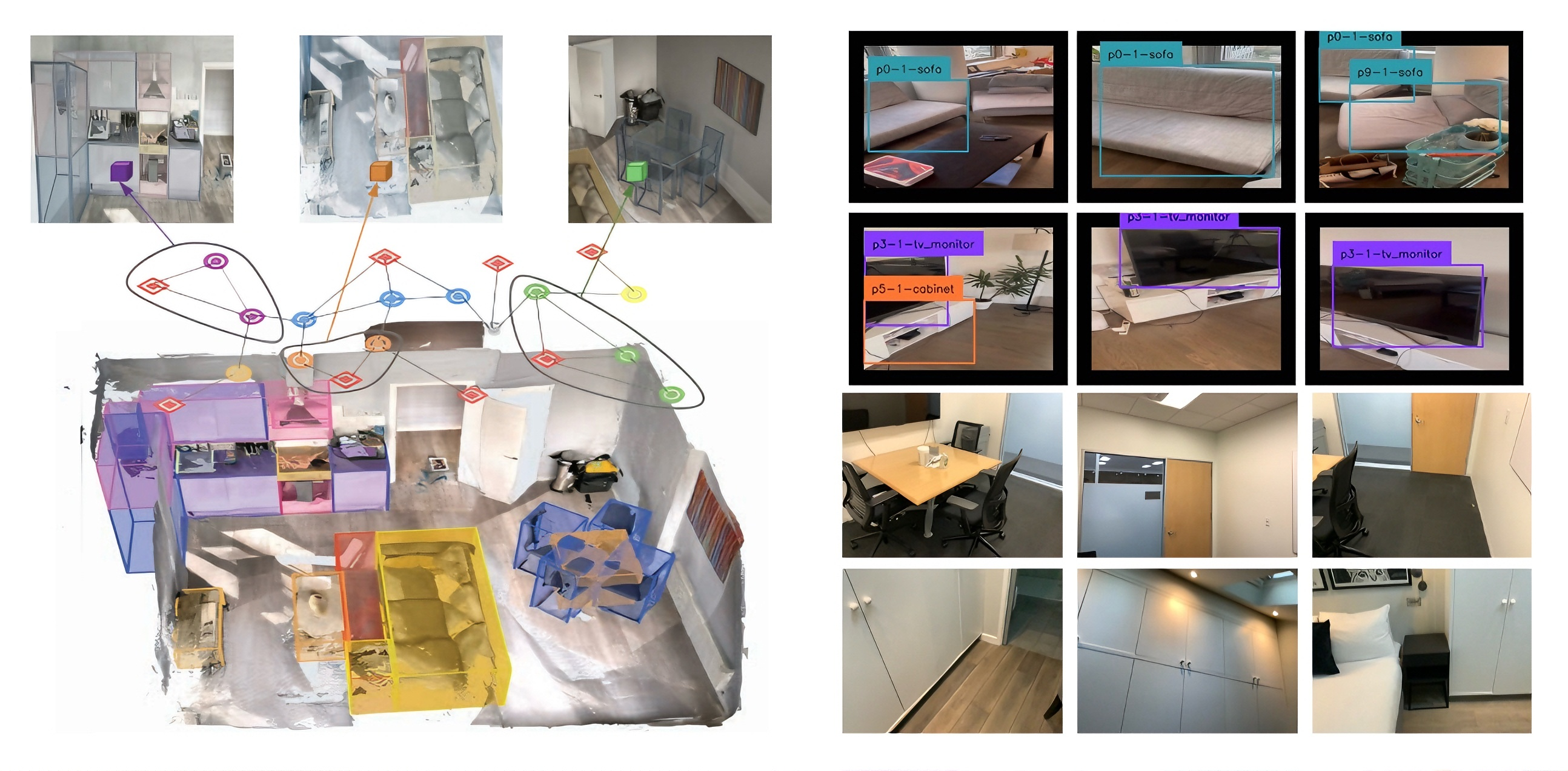}
    \caption{\textbf{Scene graph construction and cross-view consistency.} \textbf{Left:} Multi-view observations are aggregated into a hierarchical scene graph grounded in the reconstructed 3D scene. \textbf{Right:} The same object and place observed across different images, demonstrating consistent representations despite viewpoint changes.}
    \label{fig:scene_graph}
    \vspace{-0.4cm}
\end{figure}
\textbf{Scene graph.} Scene graphs provide structured representations of visual scenes by modeling objects as nodes and their semantic or spatial relationships as edges, enabling compositional reasoning for tasks such as image captioning and image retrieval \cite{krishna2017visual, xu2017scene, zellers2018neural, johnson2015image}. Recent advances improve relational modeling using transformer-based architectures, such as RelTR \cite{cong2023reltr} and SGTR \cite{li2022sgtr}, as well as open-vocabulary methods that leverage vision-language pretraining to enhance generalization beyond fixed category sets~\cite{li2024pixels}. However, conventional scene graphs are limited to 2D image representations and lack explicit spatial grounding. To address this limitation, 3D scene graphs extend the concept into spatial environments, representing scenes as topological graphs with objects, rooms, or viewpoints as nodes, constructed from 3D meshes, point clouds, or RGB-D observations~\cite{wald2020learning, rosinol2020kimera, hughes2022hydra}. These representations enable spatial reasoning but typically rely on accurate reconstruction or sequential observations. More recently, multiview-based approaches have been proposed to build scene graphs from unordered image collections without requiring explicit reconstruction. In particular, Multiview Scene Graphs (MSG)\cite{zhang2024multiview} construct scene graphs from unposed images by leveraging multiview visual correspondences to model relationships between objects and places. MSG complements existing scene graph representations, as conventional object-object relationship edges can be seamlessly integrated to enrich semantic structure, providing improved flexibility and representational depth.

\noindent\textbf{Hyperbolic representations in computer vision.} Hyperbolic representations have emerged as an effective approach for modeling hierarchical and structured data due to the exponential volume growth of hyperbolic space, which naturally captures latent semantic hierarchies. Early work by Khrulkov et al.~\cite{khrulkov2020hyperbolic} learns hyperbolic image embeddings from image-label pairs, demonstrating improved hierarchical organization compared to Euclidean embeddings. Subsequent work extends hyperbolic geometry to dense prediction tasks, such as hyperbolic image segmentation~\cite{atigh2022hyperbolic}, where hyperbolic embeddings improve the representation of semantic hierarchies between object classes. More recently, hyperbolic geometry has been incorporated into contrastive self-supervised learning frameworks to learn structured visual representations. Ermolov et al.~\cite{ermolov2022hyperbolic} introduce contrastive learning in hyperbolic space to better preserve hierarchical similarities, while Ge et al.~\cite{ge2023hyperbolic} further demonstrate the benefits of hyperbolic contrastive learning for visual representation learning. These approaches extend standard Euclidean contrastive frameworks such as SimCLR~\cite{chen2020simple} and instance discrimination~\cite{wu2018unsupervised} by replacing Euclidean similarity with hyperbolic distance, improving the modeling of semantic relationships. Beyond unimodal vision tasks, hyperbolic representations have also been applied to multimodal learning. MERU~\cite{desai2023hyperbolic} learns hyperbolic vision-language representations at scale and demonstrates strong zero-shot transfer performance across a wide range of downstream tasks, highlighting the effectiveness of hyperbolic geometry for capturing hierarchical structure in large-scale multimodal data. More recent works\cite{pal2024compositional, sinha2024learning, zheng2025hyperbolic, lin2023hyperbolic} further explore hyperbolic representation learning for vision-language alignment, retrieval, and structured scene understanding, showing consistent improvements in modeling semantic hierarchies and relational structure compared to Euclidean embeddings.

\noindent\textbf{Entailment embeddings.}
Entailment embeddings aim to model asymmetric semantic relations such as \textit{is-a} relations directly in the embedding space. Order embeddings \cite{vendrov2015order} represent hierarchical relations using partial orders in Euclidean space, enforcing that more general concepts dominate more specific ones along each dimension. This idea was later extended to non-Euclidean geometries. In particular, hyperbolic entailment cones \cite{ganea2018hyperbolic} model entailment as geodesically convex cones in hyperbolic space, naturally capturing hierarchical structures with exponential growth. For NLP and knowledge graph embedding tasks, several methods embed partially ordered data \cite{bai2021modeling, dasgupta2020improving, nguyen2017hierarchical, vilnis2018probabilistic}. These approaches demonstrate that embedding geometries with asymmetric constraints provide a principled way to represent hierarchical relations, which has inspired subsequent work on hierarchical representation learning. 

\section{Preliminaries}\label{Preliminaries}
Hyperbolic spaces are a type of Riemannian Manifolds characterized by constant negative curvature. They are fundamentally different from Euclidean spaces which are flat and have zero curvature. Hyperbolic space is ideal for representing data with an inherent hierarchical or tree-like structure, as it has properties such that volumes of subsets can grow exponentially as a function of their radius. \cite{sarkar2011low,nickel2017poincare,krioukov2010hyperbolic}

There are several models of hyperbolic geometry that define a hyperbolic plane which satisfies the axioms of a hyperbolic geometry, such as the Beltrami–Klein model which represents n-dimensional hyperbolic space in $R^n$ but distorts angles or as the Poincaré disk model that preserves angles but distorts distances\cite{cannon1997hyperbolic}.

In this context we shall only focus on the Lorentz model, which is an n-dimensional manifold representing the upper half of a two-sheeted hyperboloid in $(n+1)$-dimensional Minkowski spacetime. Figure \ref{fig:visualisation} illustrates hyperboloid manifolds and their relation to the location representation of the project-place. Every vector $\mathbf{x}\in \mathbb{R} ^{n+1}$ is represented in the form $[\mathbf{x}_{space},x_{time}]$, here $\mathbf{x}_{space}\in \mathbb{R} ^{n}$ are in space dimensions and $x_{time}\in \mathbb{R}$ lies on the hyperboloid’s axis of symmetry, also refers to time dimension. 

The Lorentz model with a constant negative curvature $-c$ is defined as the set of vectors satisfying the following condition:
\begin{equation}
    \mathcal{L} ^n=\left \{ \mathbf{x}\in \mathbb{R} ^{n+1}:\left \langle \mathbf{x},\mathbf{x} \right \rangle _{\mathcal{L}}=-\frac{1}{c} \right \}\ , \quad c>0
\end{equation}
where $\left \langle \cdot ,\cdot  \right \rangle _{\mathcal{L}}$ denotes the Lorentz inner product which is induced from Riemannian metric of the Lorentz model. For two vectors $x,y\in\mathbb{R} ^{n+1}$, it is computed as follows:
\begin{equation}
    \left \langle \mathbf{x},\mathbf{y} \right \rangle _{\mathcal{L}}=\left \langle \mathbf{x}_{space},\mathbf{y}_{space} \right \rangle-x_{time}y_{time}
\end{equation}
The Lorentzian distance is length of the geodesics in the Lorentz model, where geodesic is the shortest path between two points on the manifold. For any two points $\mathbf{x},\mathbf{y}\in\mathcal{L} ^n$, the Lorentzian distance is:
\begin{equation}
    d_\mathcal{L} (\mathbf{x},\mathbf{y})=\sqrt[]{\frac{1}{c} } cosh^{-1}(-c\left \langle \mathbf{x},\mathbf{y} \right \rangle _{\mathcal{L}}).
\end{equation}
We can map a vector $\mathbf{u}\in\mathbb{R} ^{n+1}$ from Euclidean space onto the manifold by projecting it to the tangent space $\mathcal{T}_\mathbf{z} \mathcal{L} ^n$ at some point $\mathbf{z}\in {\mathcal{L}}^n$ via orthogonal projection:
\begin{equation}
    \mathbf{v} = \mathrm{proj} _\mathbf{z}(\mathbf{u} )=\mathbf{u}+c \ \mathbf{z}\left \langle \mathbf{u},\mathbf{z} \right \rangle_{\mathcal{L}}.  
\end{equation}
Then we can use the $exponantial \ map$ to map the vector $\mathbf{v}$ from the tangent space $\mathcal{T}_\mathbf{z} \mathcal{L} ^n$ to the manifold, which is defined as:
\begin{equation}
    \mathrm{expm}_\mathbf{z}(\mathbf{v})=\mathrm{cosh}(\sqrt{c} \left \| \mathbf{v} \right \| _{\mathcal{L} })\mathbf{z} + \frac{\mathrm{sinh}(\sqrt{c} \left \| \mathbf{v} \right \| _{\mathcal{L} })}{\sqrt{c} \left \| \mathbf{v} \right \| _{\mathcal{L} }}\mathbf{v}, 
\end{equation}
where $\left \| \mathbf{v}  \right \| _{\mathcal{L} }$ is the Lorentzian norm derived from the Lorentzian inner product as $\left \| \mathbf{v}  \right \| _{\mathcal{L} }=\sqrt{\left \langle \mathbf{v} ,\mathbf{v}  \right \rangle _{\mathcal{L}}}$. For our approach, we will only consider maps where $\mathbf{z}$ is the origin of the hyperboloid, i.e., $\mathrm {O} =\left [ \mathbf{0} , \sqrt{1/c} \right ] $.

The inverse transformation $the \ logarithmic \ map$ ($\mathrm{logm} _\mathbf{z} :\mathcal{L} ^n\longrightarrow \mathcal{T}_\mathbf{z} \mathcal{L} ^n$) is obtained by inverting the closed-form expression of the exponential map along geodesics in the hyperboloid model, using the Lorentzian inner product to recover both distance and direction. It is expressed as follows:
\begin{equation}
    \mathbf{v}=\log_{\mathbf{z}}(\mathbf{x})
=\frac{\cosh^{-1}\!\bigl(-c\,\langle \mathbf{z},\mathbf{x}\rangle_{\mathcal{L}}\bigr)}
{\sqrt{\bigl(c\,\langle \mathbf{z},\mathbf{x}\rangle_{\mathcal{L}}\bigr)^2-1}}
\,\mathrm{proj}_{\mathbf{z}}(\mathbf{x}).
\end{equation}
Since Hyperbolic space is a Hadamard manifold, the Cartan–Hadamard Theorem induced that the exponential map is a global diffeomorphism and hence the logarithmic map is uniquely defined on $\mathcal{L} ^n$ \cite{do1992riemannian}.

\begin{figure}[t]
    \centering
    \includegraphics[width=\linewidth]{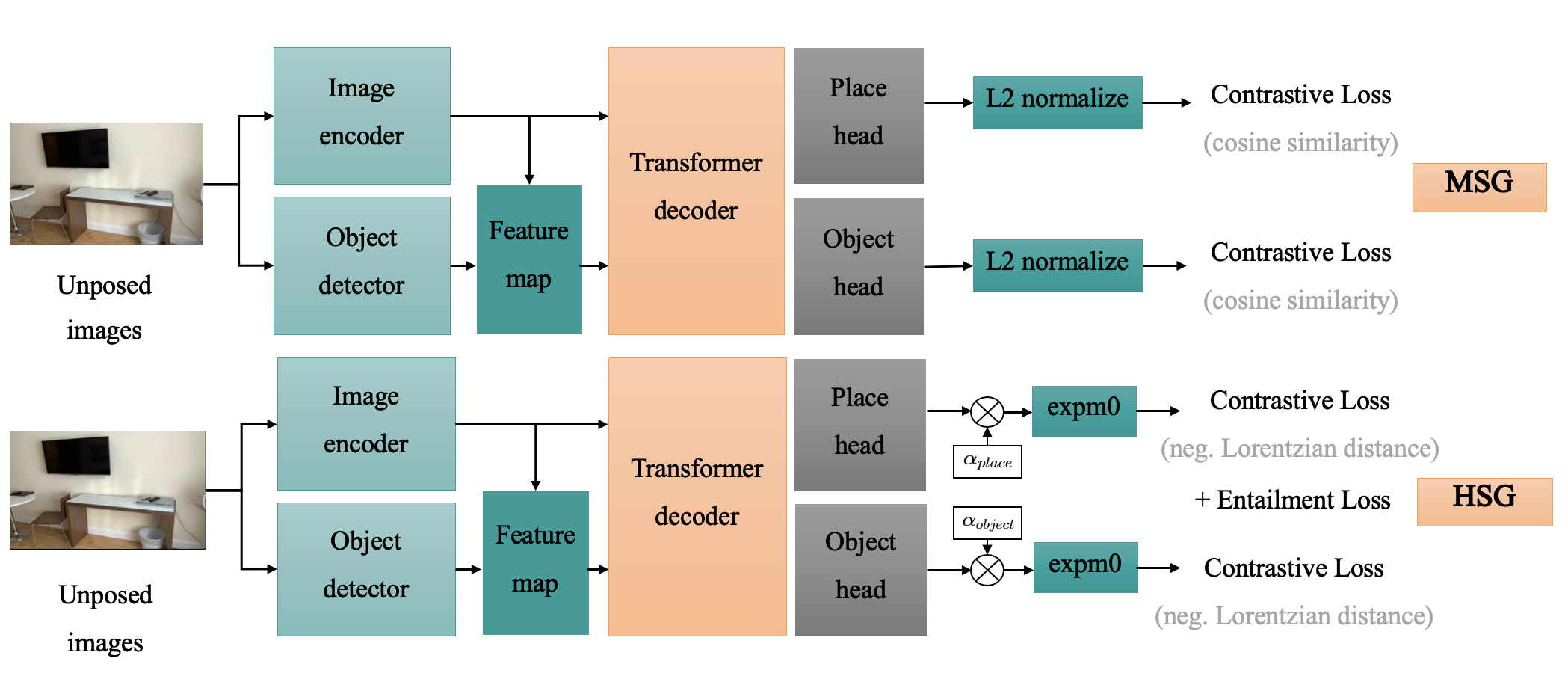}
    \caption{\textbf{HSG model design:} HSG adopts a similar architecture to MSG, but replaces L2-normalized hyperspherical embeddings and cosine similarity with Lorentz hyperboloid embeddings via the exponential map and negative Lorentzian distance, along with an additional entailment loss to enforce hierarchical structure.}
    \label{fig:model}
    \vspace{-0.4cm}
\end{figure}

\section{The Proposed Method}
In this section, we describe the pipeline that uses scene graphs as the output representation. We further introduce the role of hyperbolic deep learning in our framework, which serves as a key component of the model. Throughout this section, we use insights from hyperbolic geometry introduced in Section~\ref{Preliminaries}.

Our model design is inspired by MERU~\cite{desai2023hyperbolic} due to its simplicity and scalability. The key design choices are: (1) mapping embeddings to the Lorentz hyperboloid to enable learning in hyperbolic space, and (2) optimizing a Lorentzian contrastive objective together with an entailment loss that encourages hierarchical relationships between place and object representations, as illustrated in Figure~\ref{fig:model}.

\subsection{Hyperbolic Deep Learning}
\noindent\textbf{Projecting embeddings onto the hyperboloid.}
Let the decoder output of AoMSG after linear projection be $\mathbf{v}_{dec} \in \mathbb{R}^n$. To model hierarchical scene graph structures in hyperbolic space, we lift these Euclidean embeddings onto the Lorentz hyperboloid $\mathcal{L}^n_c \subset \mathbb{R}^{n+1}$ with curvature $-c$. We construct a vector $\mathbf{v} = [\mathbf{v}_{dec}, 0] \in \mathbb{R}^{n+1}$, which lies in the tangent space at the hyperboloid origin $\mathbf{o} = (0, \dots, 0, 1/\sqrt{c})$. Under the Lorentzian inner product $\langle \cdot, \cdot \rangle_{\mathcal{L}}$, we have $\langle \mathbf{o}, \mathbf{v} \rangle_{\mathcal{L}} = 0$, thus satisfying the tangent constraint. We parameterize only the space components of the Lorentz model, i.e., $\mathbf{v}_{dec} = \mathbf{v}_{space}$, and simplify the exponential map accordingly.

The exponential map at the origin is given by
\begin{equation}
\operatorname{Exp}_{\mathbf{o}}(\mathbf{v}) 
= \cosh(\sqrt{c}\|\mathbf{v}\|_{\mathcal{L}})\mathbf{o} 
+ \frac{\sinh(\sqrt{c}\|\mathbf{v}\|_{\mathcal{L}})}{\sqrt{c}\|\mathbf{v}\|_{\mathcal{L}}} \mathbf{v}.
\end{equation}

Since the time component of $\mathbf{v}$ is zero, its Lorentz norm reduces to the Euclidean norm of the space part, i.e., $\|\mathbf{v}\|_{\mathcal{L}} = \|\mathbf{v}_{space}\|_2$. Furthermore, the space component of the first term vanishes, leading to a simplified expression for the projected embedding:
\begin{equation}
\mathbf{x}_{space} 
= \frac{\sinh(\sqrt{c}\|\mathbf{v}_{space}\|_2)}{\sqrt{c}\|\mathbf{v}_{space}\|_2} \mathbf{v}_{space}.
\end{equation}

The corresponding time component is recovered from the hyperboloid constraint
\begin{equation}
x_{time} 
= \sqrt{\frac{1}{c} + \|\mathbf{x}_{space}\|_2^2},
\end{equation}
which guarantees that the resulting vector $\mathbf{x} = [\mathbf{x}_{space}, x_{time}]$ satisfies $\langle \mathbf{x}, \mathbf{x} \rangle_{\mathcal{L}} = -1/c$. This parameterization eliminates the need for explicit orthogonal projection onto the manifold and ensures that all embeddings lie exactly on the hyperboloid during training.

\noindent\textbf{Avoiding numerical overflow.}
The hyperbolic exponential map involves hyperbolic functions whose values grow exponentially with the tangent norm. When $\|\mathbf{v}_{space}\|_2$ becomes large, numerical overflow may occur due to the rapid growth of $\sinh(\cdot)$ and $\cosh(\cdot)$. To maintain stability, we constrain the tangent norm before applying the exponential map:
\begin{equation}
\mathbf{v}_{space} \leftarrow 
\frac{\min(\|\mathbf{v}_{space}\|_2, r_{max})}{\|\mathbf{v}_{space}\|_2} \mathbf{v}_{space},
\end{equation}
where $r_{max}$ is chosen such that $\sqrt{c} r_{max}$ remains within a numerically stable range.

The Lorentzian geodesic distance between two points $\mathbf{x}, \mathbf{y} \in \mathcal{L}^n_c$ is defined as
\begin{equation}
d_{\mathcal{L}}(\mathbf{x}, \mathbf{y}) 
= \frac{1}{\sqrt{c}} 
\operatorname{arcosh}\!\left(-c \langle \mathbf{x}, \mathbf{y} \rangle_{\mathcal{L}}\right).
\end{equation}

Since $\operatorname{arcosh}(z)$ is defined only for $z \ge 1$, we clamp the argument to $1 + \epsilon$ to avoid invalid values caused by floating-point errors during optimization.

\noindent\textbf{Contrastive learning formulation.}
We reformulate both object-level and predicate-level supervision into a hyperbolic contrastive objective, following a similar formulation to prior contrastive learning frameworks\cite{sohn2016improved, radford2021learning}. Given an anchor embedding $\mathbf{x}_i$, a positive embedding $\mathbf{x}_i^{+}$, and a set of negatives $\{\mathbf{x}_j^{-}\}$, we define similarity using the negative Lorentz distance. The resulting InfoNCE objective becomes
\begin{equation}
\mathcal{L}_{NCE}
=
- \log
\frac{
\exp\!\left(-d_{\mathcal{L}}(\mathbf{x}_i, \mathbf{x}_i^{+})/\tau\right)
}{
\exp\!\left(-d_{\mathcal{L}}(\mathbf{x}_i, \mathbf{x}_i^{+})/\tau\right)
+
\sum_j
\exp\!\left(-d_{\mathcal{L}}(\mathbf{x}_i, \mathbf{x}_j^{-})/\tau\right)
}.
\end{equation}
Here, $\tau$ denotes the temperature parameter. Compared to Euclidean similarity, Lorentz distance grows approximately linearly with radial separation but exponentially with hierarchical depth, enabling stronger discrimination between semantically distant concepts. This formulation encourages semantically aligned object–predicate pairs to lie close along geodesics while leveraging the curvature to allocate more representational capacity to higher-level abstractions.

\noindent\textbf{Entailment loss.}
Scene graphs exhibit intrinsic containment relations, where specific objects are localized within broader places. To model the place-object hierarchy, we adopt hyperbolic entailment cones in the Lorentz model $\mathbb{L}^n$ \cite{ge2023hyperbolic, le2019inferring, ganea2018hyperbolic, desai2023hyperbolic}. Given a place embedding $\mathbf{q}$ and an object embedding $\mathbf{p}$, the place defines a conical region $\mathcal{R}_{\mathbf{q}}$ that contains its valid child concepts. The half-aperture of the cone is defined as:
\begin{equation}
\omega(\mathbf{q}) 
= 
\sin^{-1}\!\left(
\frac{2K}{\sqrt{\kappa}\,\|\tilde{\mathbf{q}}\|}
\right),
\end{equation}
where $-\kappa$ is the curvature and $K$ controls values near the origin. 
More general places (closer to the origin) induce wider cones, while specific concepts produce narrower ones. To determine whether $\mathbf{p}$ lies inside the cone of $\mathbf{q}$, we compute the exterior angle
\begin{equation}
\phi(\mathbf{p}, \mathbf{q})
=
\cos^{-1}\!\left(
\frac{
p_0 + q_0 \kappa \langle \mathbf{p}, \mathbf{q} \rangle_{\mathcal{L}}
}{
\|\tilde{\mathbf{q}}\|
\sqrt{(\kappa \langle \mathbf{p}, \mathbf{q} \rangle_{\mathcal{L}})^2 - 1}
}
\right).
\end{equation}

We define the entailment loss with an aperture threshold $\eta$ as
\begin{equation}
\mathcal{L}_{ent}(\mathbf{p}, \mathbf{q})
=
\max\!\left(
0,\,
\phi(\mathbf{p}, \mathbf{q}) - \eta\,\omega(\mathbf{q})
\right),
\end{equation}
where $\eta$ scales the half-aperture to increase or decrease the effective width of the entailment cone, thereby controlling the strength of hierarchical constraints.

In general, our total loss is structured as  $\mathcal{L}_{\text{total}}=\mathcal{L}_{\text{pr}}+\mathcal{L}_{\text{obj}}+\lambda \mathcal{L}_{ent}$ averaged over each minibatch.

\subsection{Scene Graph}
\textbf{Multiview scene graph.}
Given a set of calibrated views observing the same physical scene, our goal is to construct a unified scene graph that aggregates object and relation information across viewpoints. Let $\{I^k\}_{k=1}^{K}$ denote $K$ input images with known camera poses. For each view $k$, a 2D scene graph is first constructed consisting of object nodes and predicate edges. Each node corresponds to a detected object instance with category label $c_i^k$ and visual feature $f_i^k$, while each edge encodes a directed predicate $p_{ij}^k$ between object pairs.

To integrate multi-view observations, we associate object instances across views via geometric consistency and feature similarity, as illustrated in Figure \ref{fig:scene_graph}. Specifically, let $\pi_k$ denote the projection function of view $k$. Two object instances from different views are matched if their reprojected 3D positions are consistent under camera transformations and their visual embeddings exhibit high similarity. Through this cross-view association, we construct a global node set $\mathcal{V} = \{\mathbf{v}_i\}$ representing unique physical objects in the scene. Each global node aggregates multi-view features via learnable fusion:
\begin{equation}
h_i = \operatorname{Fuse}\left(\{ f_i^k \mid i \in I^k \}\right),
\end{equation}
where $\operatorname{Fuse}(\cdot)$ denotes either attention-based or averaging aggregation.

Relations are similarly merged across views. For a pair of associated global nodes $(\mathbf{v}_i, \mathbf{v}_j)$, predicates observed from different viewpoints are accumulated to form a consolidated edge representation:
\begin{equation}
e_{ij} = \operatorname{Fuse}\left(\{ p_{ij}^k \}\right).
\end{equation}
This process yields a multiview scene graph $\mathcal{G} = (\mathcal{V}, \mathcal{E})$ that encodes consistent object identities and relations across viewpoints. Compared to single-view scene graphs, the multiview formulation reduces occlusion ambiguity and leverages complementary observations, leading to more complete and structurally coherent graph representations.

To further capture contextual dependencies, message passing is performed over $\mathcal{G}$. Let $h_i^{(t)}$ denote the node representation at layer $t$. Graph updates follow
\begin{equation}
h_i^{(t+1)} = \phi \left( h_i^{(t)}, \sum_{j \in \mathcal{N}(i)} \psi(h_i^{(t)}, h_j^{(t)}, e_{ij}) \right),
\end{equation}
where $\phi$ and $\psi$ are learnable functions and $\mathcal{N}(i)$ denotes the neighborhood of node $i$. The final node and edge embeddings are then used for object classification and predicate prediction.

\section{Experiments}
\noindent\textbf{Training details.}
We follow the original MSG training protocol~\cite{zhang2024multiview} for fair comparison. DINOv2~\cite{oquab2023dinov2} serves as the visual encoder, while the AoMSG decoder~\cite{carion2020end} generates object-level embeddings via learnable object queries and cross-attention. Training data is built from ARKitScenes~\cite{baruch2021arkitscenes}, where 3D object bounding boxes are projected into image frames using camera poses, and place labels are derived from relative pose thresholds (1\,m translation, 1\,rad rotation). Images are resized to $192\times256$, yielding 4,492 training scenes and 200 test scenes without overlap. Models are trained end-to-end using AdamW~\cite{loshchilov2017decoupled} with a learning rate of $2\times10^{-6}$, weight decay 0.01, 3-epoch linear warmup followed by cosine decay~\cite{loshchilov2016sgdr}, for 10 epochs in total. For hyperbolic models, curvature is initialized as $\texttt{curv\_init}=80.0$ and optimized as a learnable parameter, with loss weight $\lambda=20$.

\noindent\textbf{Evaluation metrics.}
We follow the evaluation protocol of Multiview Scene Graph (MSG)~\cite{zhang2024multiview} and compare predicted scene graphs with ground truth using IoU between adjacency matrices. We report Place-Place IoU (PP IoU) for place connectivity and Place-Object IoU (PO IoU) for cross-view object associations. We further compute generalized IoU (GIoU) for object bounding boxes~\cite{rezatofighi2019generalized}. These metrics evaluate structural alignment between predicted and ground-truth graphs and reflect the quality of the learned spatial representations~\cite{ristani2016performance}.
\begin{figure}[htbp]
\centering
\begin{minipage}[t]{0.46\textwidth}
    \vspace{0pt}
    \centering
    \caption{\textbf{Different curvature initializations:} Curvature initialization strongly impacts retrieval and graph-level performance, values that are too small or large cause metric collapse due to limited hierarchical capacity or instability. Performance peaks at \texttt{curv\_init} \(= 80\), where hyperbolic space balances hierarchical expressiveness and training stability.}
    \label{fig:curv_init}
    \includegraphics[width=\linewidth]{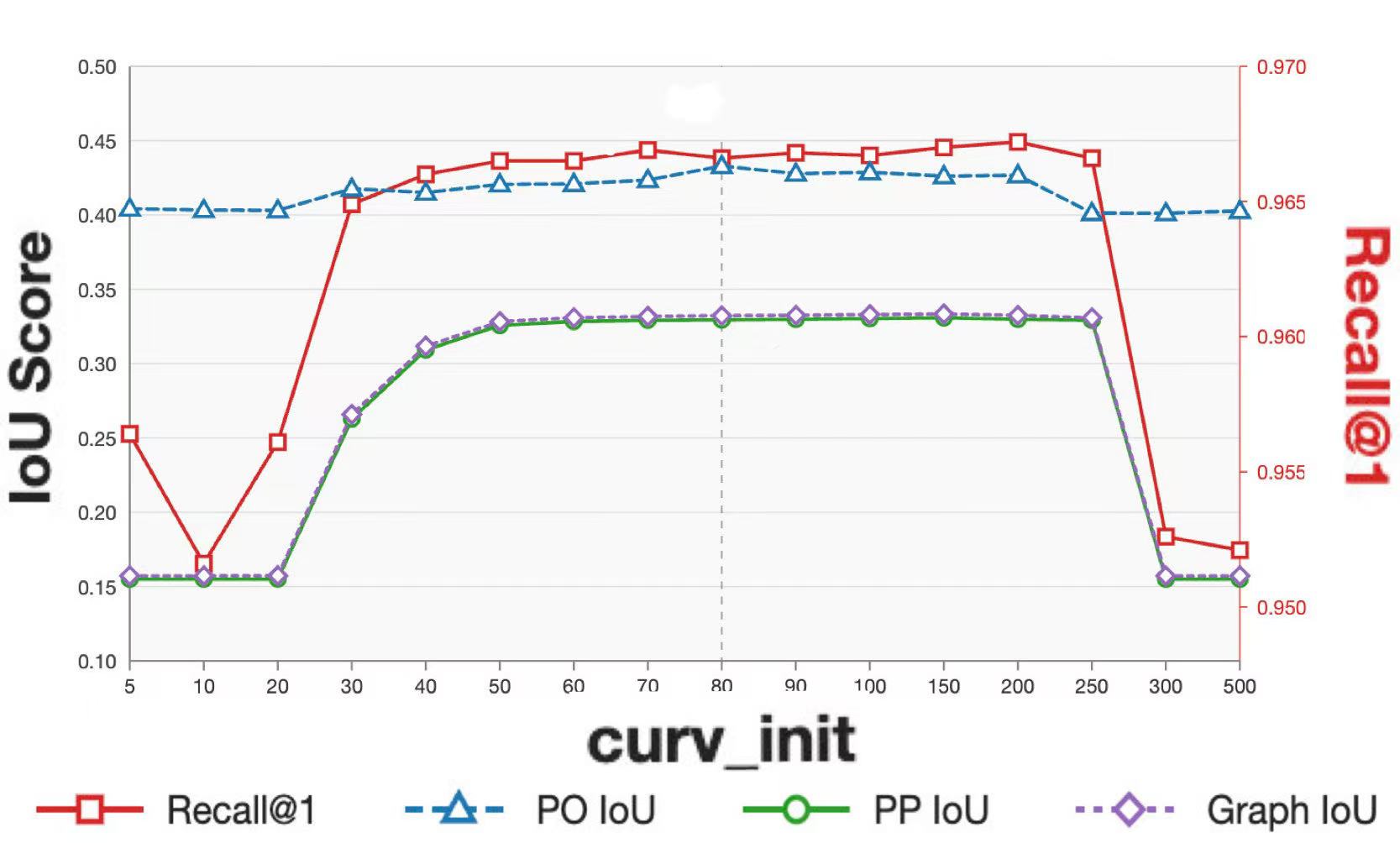}
\end{minipage}
\hfill
\begin{minipage}[t]{0.52\textwidth}
    \vspace{0pt}
    \centering
    \vspace{-1em}
    \captionof{table}{\textbf{Main results:} Our model uses DINOv2~\cite{oquab2023dinov2} as the backbone. SepMSG-Direct, SepMSG-Linear, and SepMSG-MLP are baselines that project frozen backbone features into the MSG embedding space via direct use, linear projection, and a small MLP, respectively. AoMSG-1, AoMSG-2, and AoMSG-4 denote AoMSG models with 1, 2, and 4 Transformer decoder layers. Results show that HSG outperforms nearly all baselines on PP-IoU and Graph IoU.}
    \label{Main results}
    \resizebox{\linewidth}{!}{%
    \begin{tabular}{lcccc}
    \toprule
    \multirow{2}{*}{\textbf{Method}} & \multicolumn{4}{c}{\textbf{Metric}} \\
    \cmidrule(lr){2-5}
     & Recall@1 & PP IoU & PO IoU & Graph IoU \\
    \midrule
    SepMSG-Linear\cite{zhang2024multiview}      & 97.30 & 20.02 & 55.67 & 20.48 \\
    SepMSG-MLP\cite{zhang2024multiview}         & 97.39 & 20.51 & 58.63 & 21.01 \\
    SepMSG-Direct\cite{zhang2024multiview}      & 97.71 & 33.19 & 48.58 & 33.67 \\
    AoMSG-2\cite{zhang2024multiview}            & 98.50 & 21.02 & 49.18 & 21.39 \\
    AoMSG-1\cite{zhang2024multiview}            & 98.47 & 24.87 & 55.50 & 25.37 \\
    AoMSG-B-4\cite{zhang2024multiview}          & 98.61 & 17.87 & 44.46 & 18.15 \\
    AoMSG-4 (baseline)\cite{zhang2024multiview} & 98.33 & 21.02 & 42.79 & 21.32 \\
    \midrule
    \rowcolor{gray!30}
    HSG (Ours)
    & 98.39 & 33.17 & 45.52 & 33.51 \\
    \bottomrule
    \end{tabular}}
\end{minipage}
\end{figure}

\subsection{Results}
\noindent\textbf{Main results.}
Table \ref{Main results} compares HSG with all baselines across four metrics. Overall, HSG achieves the best or near-best performance, validating the effectiveness of projecting decoder embeddings into hyperbolic space for modeling hierarchical entailment. For place retrieval, HSG attains a Recall@1 of \textbf{98.39}, remaining competitive with the strongest baseline (AoMSG-B-4, 98.61) while maintaining substantially stronger graph quality. The largest gains appear in graph-level metrics: HSG achieves a PP IoU of \textbf{33.17} and the highest Graph IoU of \textbf{33.51}, outperforming the best AoMSG variant (25.37) by \textbf{+8.14}. Although SepMSG-Linear attains a higher PO IoU (55.67), it performs markedly worse on PP IoU and Graph IoU, revealing a trade-off that HSG avoids. HSG PO‑IoU advantage is limited, as this metric relies more on accurate object-level association and detection quality than on global hierarchy.In general, HSG delivers a balanced and consistently strong performance across retrieval and structural metrics, confirming that hyperbolic projection better preserves hierarchical and compositional structure in scene graphs.

\begin{table}[t]
\centering
\caption{\textbf{Comparison of Different Projector Dimensions:} We compare HSG with AoMSG-4 and SepMSG-Linear with different projector dimensions, all using DINOv2-base\cite{oquab2023dinov2} as the backbone, as it has been shown to be important for performance in self-supervised representation learning.\cite{bardes2021vicreg, bordes2023towards, chen2020simple}}
\resizebox{\linewidth}{!}{
\begin{tabular}{c|ccc|ccc|ccc}
\toprule
\multirow{2}{*}{\textbf{Projector dimension}}
& \multicolumn{3}{c|}{\textbf{HSG (Ours)}}
& \multicolumn{3}{c|}{\textbf{AoMSG-4\cite{zhang2024multiview}}}
& \multicolumn{3}{c}{\textbf{SepMSG-Linear\cite{zhang2024multiview}}} \\
\cmidrule(lr){2-4} \cmidrule(lr){5-7} \cmidrule(lr){8-10}
& Recall@1 & PP IoU & PO IoU
& Recall@1 & PP IoU & PO IoU
& Recall@1 & PP IoU & PO IoU \\
\midrule
512  
& 96.7 & 32.0 & 43.2
& 98.4 & 20.1 & 46.2
& 98.0 & 19.5 & 57.5 \\

1024 
& 98.4 & 33.2 & 45.5
& 98.3 & 21.0 & 42.8
& 97.2 & 19.7 & 57.0 \\

2048 
& 96.9 & 29.5 & 42.9
& 97.8 & 22.3 & 42.7
& 97.6 & 20.0 & 58.6 \\

\bottomrule
\end{tabular}
}
\vspace{-0.4cm}
\label{projector_dim}
\end{table}
\noindent\textbf{Projector dimensions.}
Table \ref{projector_dim} compares projector dimensions across HSG, AoMSG-4, and SepMSG-Linear. HSG achieves the best performance at \textbf{1024}, while 512 under-parameterises the projection and 2048 reduces PP IoU to 29.5. In contrast, AoMSG-4 and SepMSG-Linear show only marginal variation across dimensions. This indicates that hyperbolic projection quality is more sensitive to projector capacity than Euclidean baselines, and we therefore adopt \textbf{1024} for all experiments.

\noindent\textbf{Choices of backbones.}
Table \ref{backbones} compares HSG across seven encoder backbones.
\textbf{DINOv2-Base} achieves the best overall performance\cite{el2024probing} and is adopted as the default. Transformer-based models generally outperform CNN-based ones on graph-level metrics. ConvNeXt-Base and DINOv2-Large obtain identical PP IoU (32.94) and Graph IoU (33.22) but remain below DINOv2-Base, suggesting that large-scale self-supervised pretraining benefits hyperbolic scene graph learning. Notably, DINOv2-Small achieves high Recall@1 yet poor graph metrics, indicating that retrieval performance alone is insufficient for evaluating backbone suitability. Overall, large-scale self-supervised pretraining proves essential for effective scene graph construction in hyperbolic space.

\noindent\textbf{Effect of curvature initializations.} 
Figure \ref{fig:curv_init} shows the impact of curvature initialization on all metrics. Both overly small and large \texttt{curv\_init} values degrade performance, with a clear optimum around \textbf{80}. 
When \texttt{curv\_init} $\leq$ 20, graph metrics collapse and Recall@1 drops, as the geometry approaches Euclidean space and loses hierarchical capacity. 
\begin{table}[t]
\centering
\begin{minipage}{0.5\textwidth}  %
\centering
\caption{\textbf{Comparison of Different Encoder Backbones:} We report results from base models for ConvNeXt\cite{liu2022convnet}, ViT\cite{dosovitskiy2020image}, ResNet\cite{he2016deep} and DINOv2\cite{oquab2023dinov2}.}
\resizebox{\linewidth}{!}{\begin{tabular}{lcccc}
\toprule
\multirow{2}{*}{\textbf{Encoder backbone}} & \multicolumn{4}{c}{\textbf{Metric}} \\
\cmidrule(lr){2-5}
 & Recall@1 & PP IoU & PO IoU & Graph IoU \\
\midrule
ConvNeXt-Tiny\cite{liu2022convnet}   & 97.63 & 20.85 & 40.15 & 21.13 \\
ConvNeXt-Base\cite{liu2022convnet}   & 96.67 & 32.94 & 42.94 & 33.22 \\
ViT-Base\cite{dosovitskiy2020image}        & 96.29 & 18.63 & 38.36 & 18.89 \\
ResNet-18\cite{he2016deep}       & 95.02 & 23.08 & 34.97 & 23.64 \\
DINOv2-Small\cite{oquab2023dinov2}    & 98.08 & 16.04 & 41.10 & 16.29 \\
DINOv2-Large\cite{oquab2023dinov2}    & 96.67 & 32.94 & 42.94 & 33.22 \\
\rowcolor{gray!30}
DINOv2-Base\cite{oquab2023dinov2}     & 98.39 & 33.17 & 45.52 & 33.51 \\
\bottomrule
\label{backbones}
\end{tabular}}
\end{minipage}
\hfill
\begin{minipage}{0.45\textwidth}  %
\centering
\caption{\textbf{Ablations on HSG:} We ablate three design choices for HSG, with DINOv2\cite{oquab2023dinov2} and ConvNext\cite{liu2022convnet} as backbones.}
\resizebox{\linewidth}{!}{\begin{tabular}{lccc}
\toprule
 & Recall@1 & PP IoU & PO IoU \\
\midrule
\rowcolor{gray!30}
HSG DINOv2-Base 
& 98.4 & 33.2 & 45.5 \\
1.~~\textit{InfoNCE only}
& 96.4 & 21.5 & 40.6 \\
2.~~\textit{no entailment loss}
& 98.3 & 33.5 & 44.9 \\
3.~~\textit{fixed c=1}
& 97.9 & 15.5 & 40.6 \\
\midrule
\rowcolor{gray!30}
HSG ConvNeXt-Base 
& 96.7 & 32.9 & 42.9 \\
1.~~\textit{InfoNCE only}
& 97.9 & 15.8 & 38.3 \\
2.~~\textit{no entailment loss}
& 98.4 & 32.4 & 39.4 \\
3.~~\textit{fixed c=1}
& 98.1 & 15.5 & 39.3 \\
\bottomrule
\end{tabular}}
\end{minipage}
\vspace{-0.4cm}
\end{table}

\noindent Performance stabilizes and peaks between 30 and 250, reaching the best values near \textbf{80}, where curvature is sufficient to model entailment while remaining well-conditioned. For curv $\geq$ 300, metrics collapse again due to numerical instability and boundary concentration effects. Excessively high curvature causes the manifold to become numerically ill conditioned, and the exponential map concentrates embeddings near the boundary of the hyperboloid, leading to vanishing gradients and loss of discriminative structure.
We therefore adopt \texttt{curv\_init} $=$ \textbf{80} as the default, balancing geometric expressiveness and training stability.

\noindent\textbf{Fixed curvature parameter c=1.}
Fixing the curvature to $c=1$ causes the entailment losses to collapse to zero, rendering hierarchical constraints ineffective. PP IoU drops sharply for DINOv2-Base and ConvNeXt-Base with PO IoU also decreasing, while Recall@1 remains high, indicating that instance discrimination is preserved but hierarchical structure is lost. The results underscore the importance of proper curvature configuration for hierarchical representation learning, although few prior works explicitly learn curvature~\cite{atigh2022hyperbolic, khrulkov2020hyperbolic, nickel2018learning}.

\noindent\textbf{No entailment loss.}
We further remove the entailment loss while keeping the model in hyperbolic space. Compared with the full HSG model, the performance difference is moderate but consistent. For DINOv2-Base and ConvNeXt-Base, PO IoU decrease mildly. This indicates that the entailment loss contributes to better modeling of place-object inclusion relationships, leading to improved object alignment and relational consistency. Overall, while hyperbolic learning provides the major structural benefit, the entailment loss further refines the hierarchical representation and improves graph quality.

\noindent\textbf{Original Euclidean contrastive loss.}
Replacing the hyperbolic objective with Euclidean InfoNCE severely degrades graph-level performance. When DINOv2 and ConvNeXt are used as encoder backbones, both PP IoU and Graph IoU decrease substantially. Despite relatively high Recall@1, the sharp decline in graph metrics confirms that Euclidean contrastive learning fails to preserve hierarchical structure, underscoring the necessity of hyperbolic representation learning.

\noindent\textbf{Effect of aperture thresholds.}
Varying the aperture threshold from 1.0 to 0.01 yields nearly identical results across all evaluation metrics, indicating that the method is robust to the aperture threshold and that hyperbolic entailment provides stable hierarchical supervision without sensitive hyperparameter tuning.

\section{Qualitative Analysis}

\begin{figure*}[t]
    \centering
    
    \begin{subfigure}{0.48\textwidth}
        \centering
        \includegraphics[width=\linewidth]{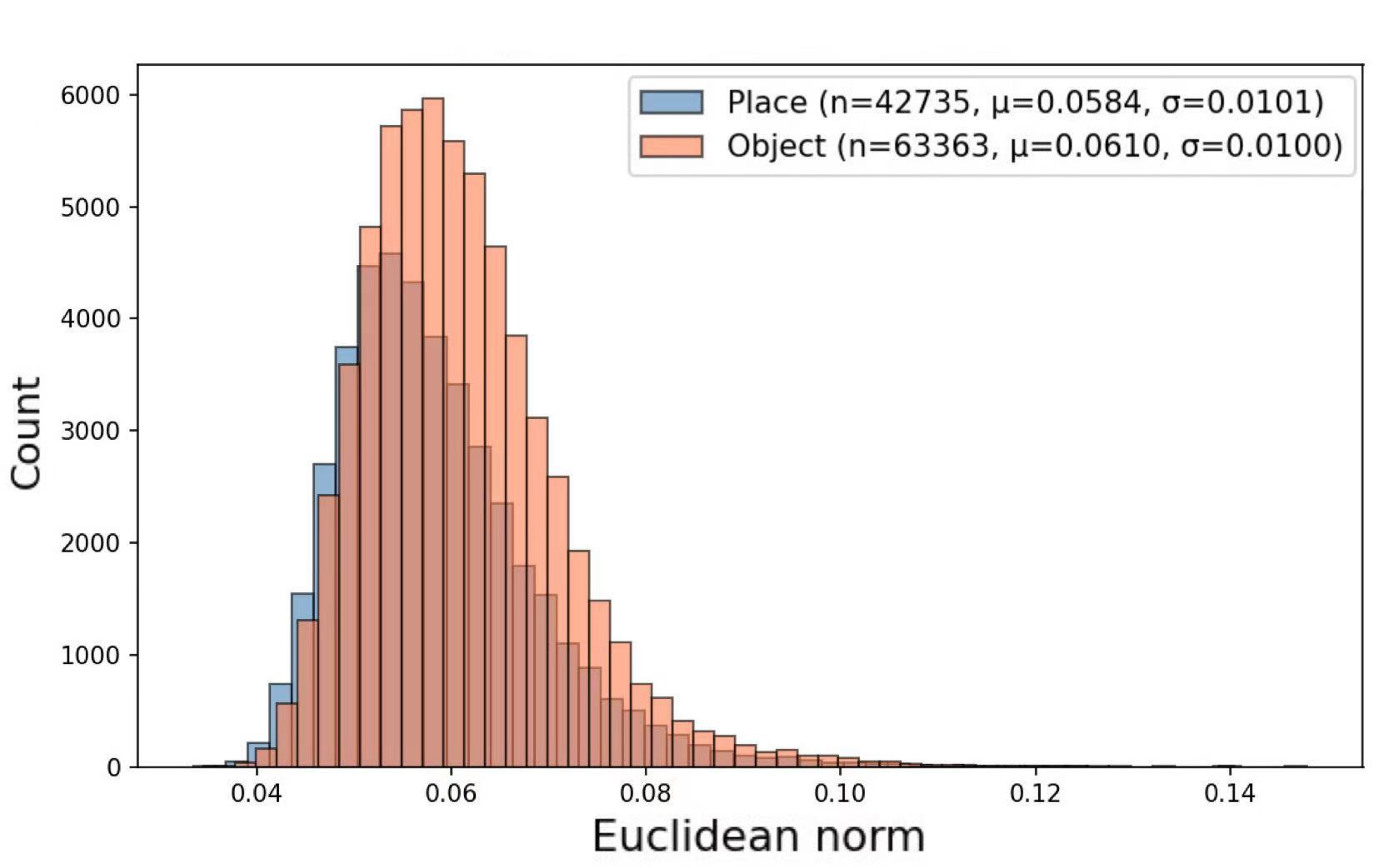}
        \caption{HSG}
        \label{fig:HSG}
    \end{subfigure}
    \hfill
    \begin{subfigure}{0.48\textwidth}
        \centering
        \includegraphics[width=\linewidth]{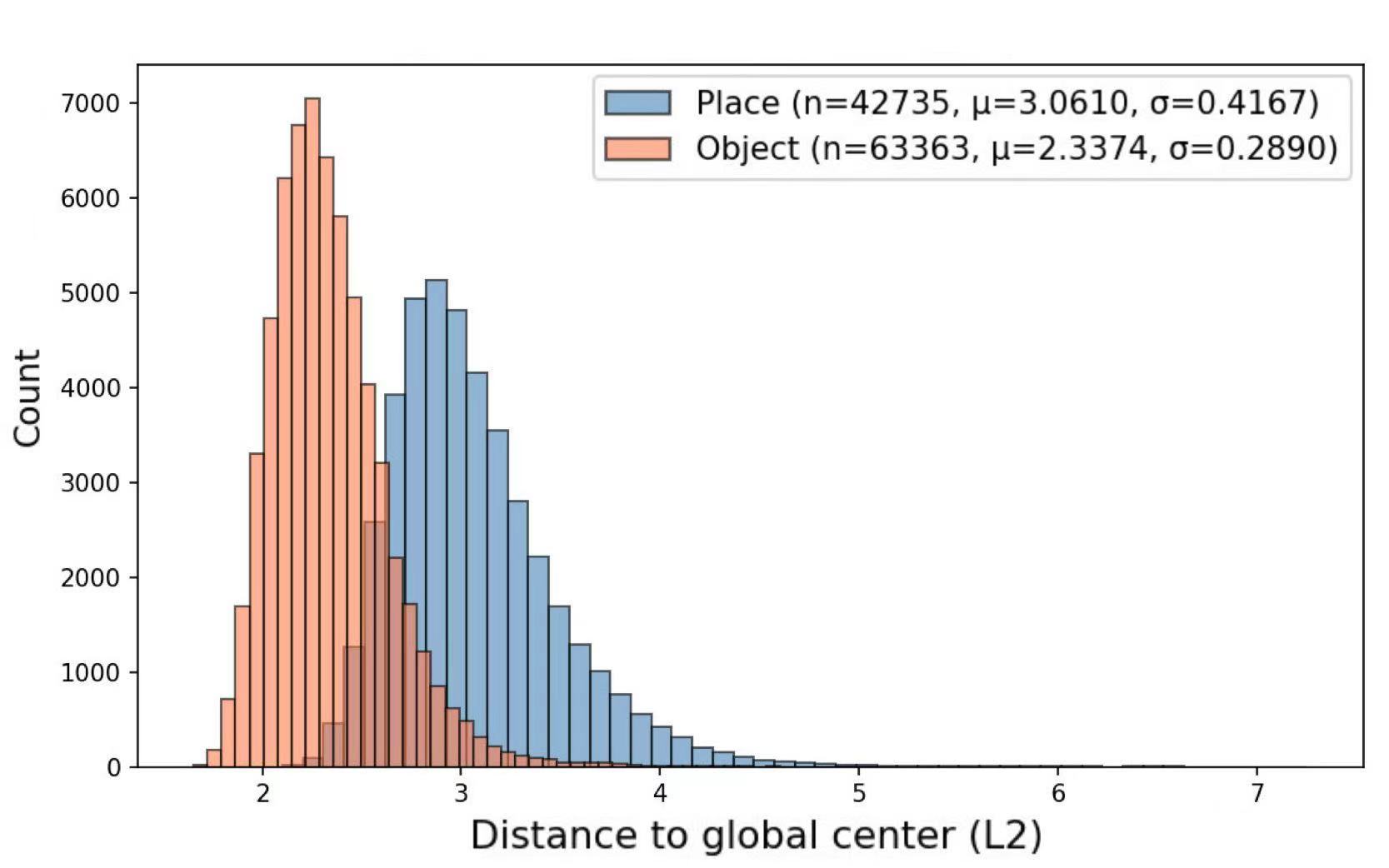}
        \caption{AoMSG}
        \label{fig:AoMSG}
    \end{subfigure}
    \hfill
    \begin{subfigure}{0.48\textwidth}
        \centering
        \includegraphics[width=\linewidth]{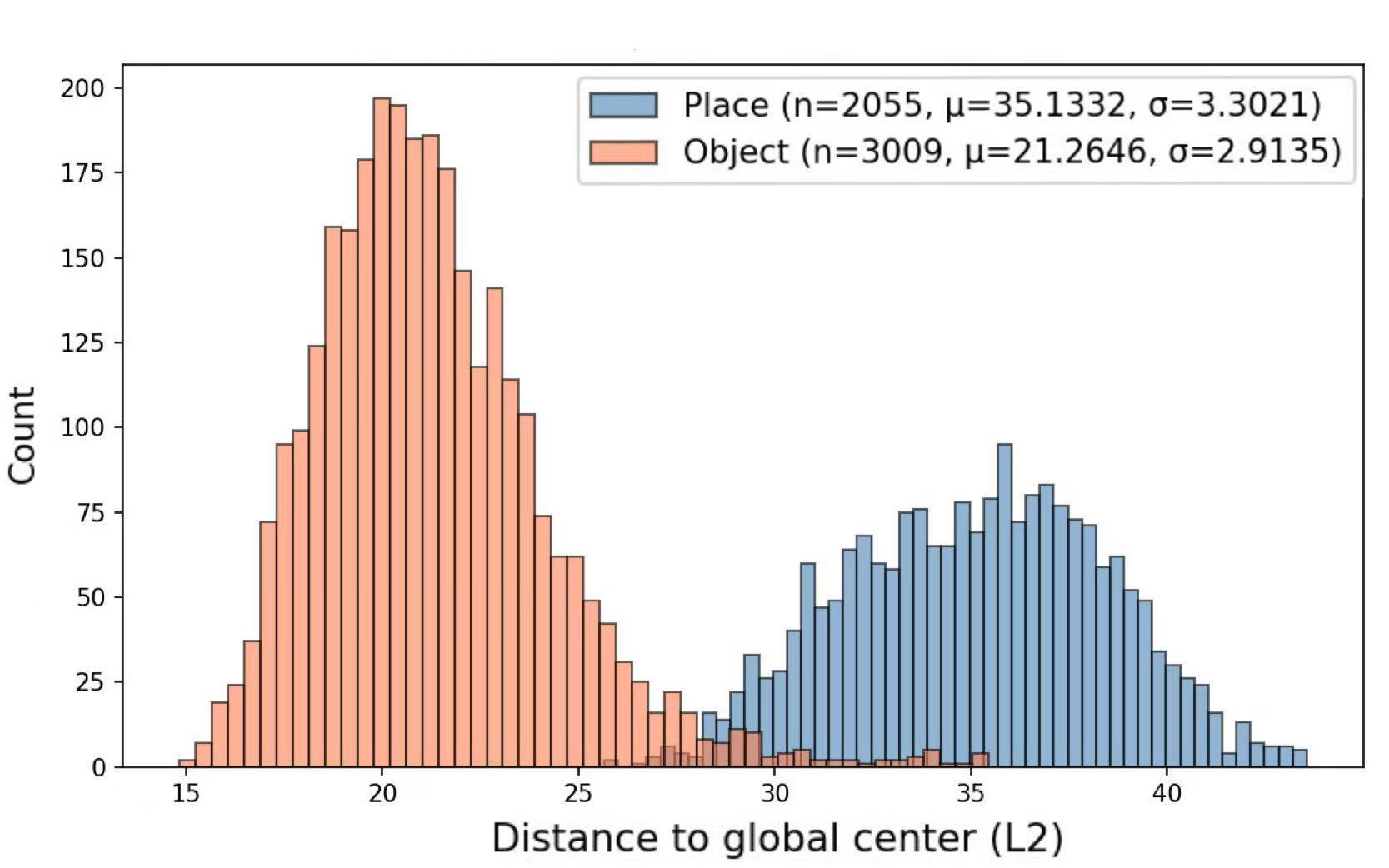}
        \caption{SepMSG-Liner}
        \label{fig:SepMSG-Liner}
    \end{subfigure}
    \hfill
    \begin{subfigure}{0.48\textwidth}
        \centering
        \includegraphics[width=\linewidth]{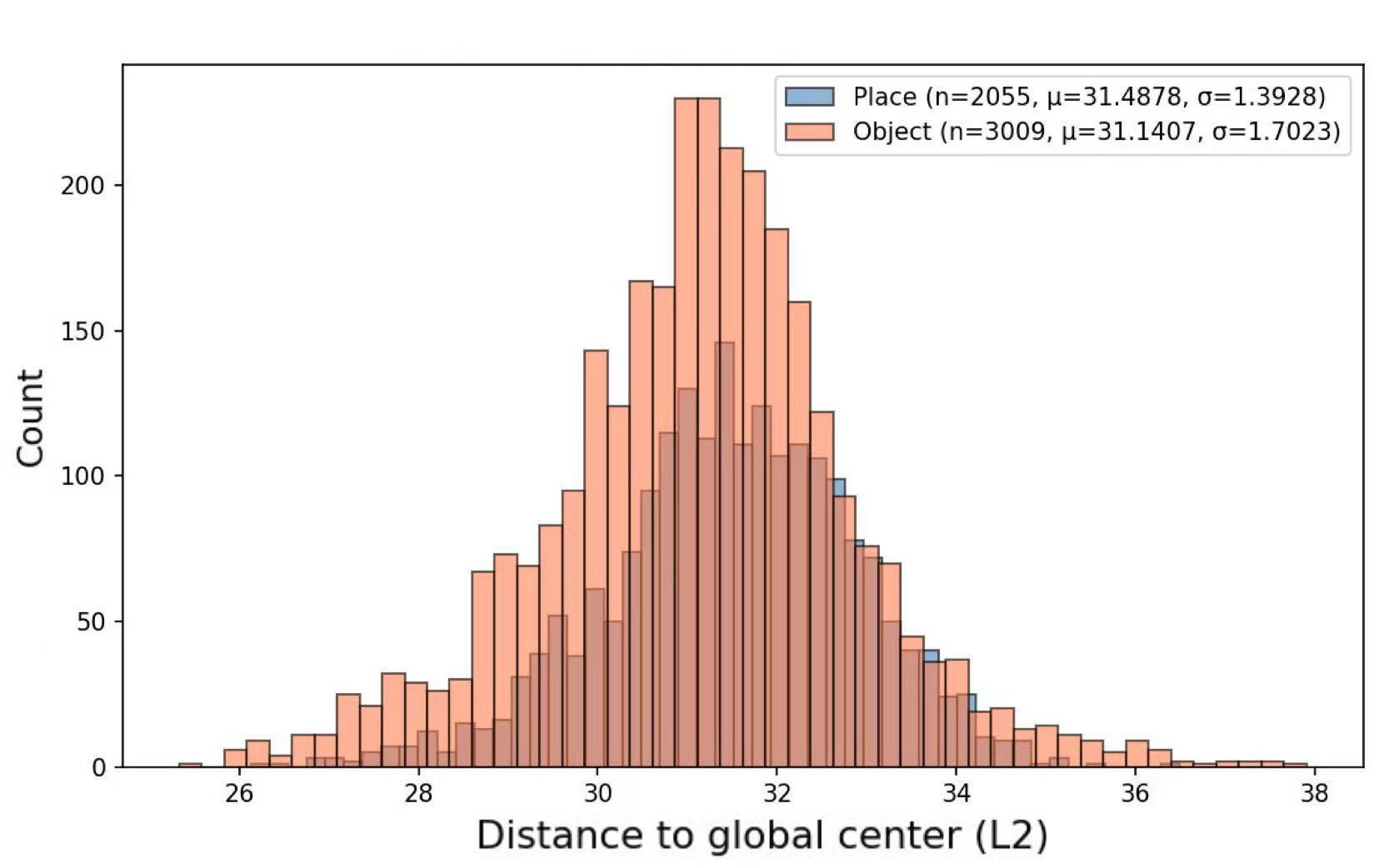}
        \caption{SepMSG-MLP}
        \label{fig:SepMSG-MLP}
    \end{subfigure}

    \caption{\textbf{Distribution of embedding distances from [ROOT]:} We embed training images using trained HSG, AoMSG and SepMSG, where for HSG we calculate Lorentzian norm $d(\mathbf{v})=||\mathbf{v}_{space}||$ and $L^2$ norm for AoMSG and SepMSG. Our hyperbolic representation more effectively preserves hierarchical relationships, with place nodes embedded closer to the root node in accordance with the hierarchy, whereas Euclidean baselines show no evident hierarchical structure.}
    \label{fig:qualitative}
    \vspace{-0.4cm}
\end{figure*}

\textbf{[ROOT] embedding.}
We interpret the visual-semantic hierarchy as a tree structure with a root node representing the most generic concept, denoted as \texttt{[ROOT]}, which serves as a global reference for measuring semantic specificity via distance\cite{dijkstra2022note}. Following MERU~\cite{desai2023hyperbolic}, in HSG the \texttt{[ROOT]} is naturally defined as the origin of the Lorentz hyperboloid, where distance to the origin reflects the level of abstraction. In contrast, for AoMSG and other Euclidean models without a canonical root, we approximate \texttt{[ROOT]} as the global average of all training embeddings and compute $\ell_2$ distance to provide a consistent reference for hierarchical comparison.
 
\noindent\textbf{Embedding distances from [ROOT].} To qualitatively evaluate hierarchical structure, we visualize embedding distances to the root for HSG, AoMSG, and SepMSG (Fig.~\ref{fig:qualitative}). In hyperbolic space, root distance reflects generality, with more abstract concepts expected to lie closer to the origin \cite{nickel2017poincare, ganea2018hyperbolic}. Under our formulation, \textit{place} is more general than \textit{object} and should therefore be embedded closer to the root. As shown in Fig.~\ref{fig:qualitative}(a), HSG exhibits a clear separation: \textit{place} embeddings cluster at smaller root distances, while \textit{object} embeddings lie farther away, consistent with the intended entailment structure. In contrast, AoMSG and SepMSG (Fig.~\ref{fig:qualitative}(b),(c),(d)) show weak or inconsistent ordering, indicating limited hierarchical organization. These results qualitatively confirm that HSG more effectively captures semantic hierarchy through hyperbolic geometry.

\section{Limitations and Future Work}
While HSG significantly enhances hierarchical place--object representations, several avenues remain for further improvement. First, adaptive or multi-stage curvature optimization~\cite{gu2018learning, sala2018representation, nickel2017poincare, bachmann2020constant} could more effectively accommodate non-uniform geometric structures, potentially improving both the expressive capacity of embeddings and training stability. Second, the overall performance is closely tied to the quality of the underlying encoders; integrating stronger foundation models such as DINOv3~\cite{simeoni2025dinov3}, employing more accurate open-vocabulary detectors like GroundingDINO~\cite{liu2024grounding}, or leveraging large-scale and temporal scene graph frameworks~\cite{rosinol2020kimera, johnson2015image, yang2018graph} may further enhance robustness, generalization, and scalability. Finally, exploring joint optimization with downstream tasks or incorporating multimodal cues could unlock additional benefits. Overall, hyperbolic scene graph representations offer a promising and flexible basis for scalable hierarchical visual understanding and structured reasoning in complex scenes.

\section{Conclusion}
In this paper, we propose Hyperbolic Scene Graph (HSG), a framework that learns scene graph representations in hyperbolic space to explicitly capture hierarchical place--object relationships. By combining hyperbolic contrastive learning with an entailment-aware objective, HSG produces structured and semantically consistent embeddings while remaining compatible with existing MSG pipelines. Experimental results show that HSG improves hierarchical structure quality, achieves competitive performance compared to Euclidean baselines, and extends the capability of hyperbolic representation learning to scene understanding and recognition tasks. Moreover, HSG provides a flexible foundation for incorporating additional modalities or downstream tasks, highlighting the broader potential of hyperbolic embeddings for structured visual reasoning. We hope this work demonstrates the effectiveness of hyperbolic representation learning for scene graph modeling and inspires further exploration of geometry-aware visual representations.

\bibliographystyle{splncs04}
\bibliography{main}

\clearpage
\appendix

\section{Hyperparameter setting}
\begin{table}[h]
\centering
\caption{Hyperparameters used in the HSG main experiments.}
\label{tab:hyperparameters}
\begin{tabular}{cc}
\toprule
\textbf{Hyperparameter} & \textbf{Value/Range} \\
\midrule
Original image size & $192 \times 256$ \\
Input image size & $224 \times 224$ \\
Batch size & 32 \\
Scenes per training batch & 2 \\
Images per scene per batch & 64 \\
Learning rate & $2e^{-6}$ \\
Epochs & 10 \\
Optimizer & AdamW \\
Scheduler & \texttt{linear\_warmup\_cosine} \\
Warm-up epochs & 3 \\
Weight decay & 0.01 \\
HSG layers & 4 \\
HSG patch size & 14 \\
HSG hidden dim & 384 \\
Projector head dim & 1024 \\
$curv_{init}$ & 80 \\
$L_{place}$ Place Loss Function & InfoNCE w/ $\tau=0.1$ \\
$L_{object}$ Object Loss Function & InfoNCE w/ $\tau=0.1$ \\
Loss ratio $L_{place}$ : $L_{object}$ : $L_{entail}$ & 1 : 1 : 20 \\
Place Similarity Threshold & 0.3 \\
Object Similarity Threshold & 0.2 \\
\bottomrule
\end{tabular}
\end{table}

\label{sec:hyper_setting}

\section{Details of evaluation metrics}

\subsection{IoU Between Two Adjacency Matrices}
Let $A \in \{0,1\}^{n \times n}$ denote the ground-truth adjacency matrix of a multiview scene graph and 
$\hat{A} \in \{0,1\}^{m \times m}$ denote the predicted adjacency matrix.
After aligning the vertices between the two graphs, we compute the intersection-over-union (IoU) between the two adjacency matrices.
Let the element-wise intersection and union be defined as

\begin{equation}
I = A \land \hat{A},
\end{equation}

\begin{equation}
U = A \lor \hat{A}.
\end{equation}

The graph IoU is then defined as

\begin{equation}
\mathrm{IoU}(A,\hat{A}) =
\frac{\sum_{i,j} I_{ij}}
{\sum_{i,j} U_{ij}} .
\end{equation}

From a graph prediction perspective, this IoU evaluates binary edge prediction between the ground-truth and predicted graphs.
Let TP, FP and FN denote the number of true-positive, false-positive and false-negative edges respectively.
The IoU can equivalently be expressed as

\begin{equation}
\mathrm{IoU} =
\frac{\mathrm{TP}}
{\mathrm{TP} + \mathrm{FP} + \mathrm{FN}} .
\end{equation}

This definition naturally penalizes additional edges that appear only in the predicted graph or missing edges that appear only in the ground-truth graph.

\subsection{Place-Place and Place-Object IoU.}
The adjacency matrix can be decomposed into two subgraphs:
\begin{itemize}
\item place-place edges $A^{pp}$
\item place-object edges $A^{po}$
\end{itemize}
We therefore evaluate them separately as

\begin{equation}
\mathrm{IoU}_{pp} =
\frac{|A^{pp} \cap \hat{A}^{pp}|}
{|A^{pp} \cup \hat{A}^{pp}|}
\end{equation}

and

\begin{equation}
\mathrm{IoU}_{po} =
\frac{|A^{po} \cap \hat{A}^{po}|}
{|A^{po} \cup \hat{A}^{po}|}.
\end{equation}

The final graph IoU evaluates the alignment of the entire multiview scene graph.

\subsubsection{Object Truth-to-Result Matching.}
To evaluate the place-object edges, we must first establish correspondences
between ground-truth objects and predicted objects.

Let $o_i$ denote a ground-truth object and $\hat{o}_j$ denote a predicted object.
For each frame $t$, let $g_i^t$ denote the ground-truth detection of object $o_i$, and $\hat{g}_j^t$ denote the predicted detection of object $\hat{o}_j$.
We define indicator functions to represent the existence of the object in frame $t$:

\begin{equation}
\mathbb{1}(g_i^t)
=
\begin{cases}
1 & \text{if } o_i \text{ exists in frame } t \\
0 & \text{otherwise}
\end{cases}
\end{equation}

\begin{equation}
\mathbb{1}(\hat{g}_j^t)
=
\begin{cases}
1 & \text{if } \hat{o}_j \text{ exists in frame } t \\
0 & \text{otherwise}
\end{cases}
\end{equation}

The accumulated generalized IoU score between objects $o_i$ and $\hat{o}_j$
is defined as

\begin{equation}
S_{ij}
=
\sum_t
\mathrm{GIoU}(g_i^t, \hat{g}_j^t).
\end{equation}

The score is accumulated across all frames where the two objects appear.
A one-to-one object matching is then obtained by maximizing the accumulated score:

\begin{equation}
\pi =
\arg\max_{\pi}
\sum_i S_{i,\pi(i)} .
\end{equation}

The resulting permutation $\pi$ aligns the predicted objects with the ground-truth objects, allowing the place-object adjacency matrices to be reordered and evaluated consistently. The predicted object nodes are reordered according to the optimal matching permutation $\pi$ before computing the PO IoU.

\section{Entailment Loss Derivations}

We derive the geometric formulation of the entailment loss used in our model. 
Let $x_b$ be a point in the Poincaré ball. Following the formulation of hyperbolic entailment cones, the half-aperture of the cone centered at $x_b$ is defined as

\begin{equation}
\mathrm{aper}_b(x_b)=
\sin^{-1}\!\left(
K \frac{1-c\|x_b\|^2}{\sqrt{c}\|x_b\|}
\right),
\end{equation}

where $c>0$ denotes the magnitude of curvature and $K$ is a constant controlling the cone scale. 
The Poincaré ball model and the Lorentz hyperboloid model are isometric representations of hyperbolic space, therefore a point $x_b$ in the Poincaré ball can be mapped to a point $x_h$ on the hyperboloid through the differentiable transformation

\begin{equation}
x_h=\frac{2x_b}{1-c\|x_b\|^2}.
\end{equation}

Since the cone aperture should be invariant across equivalent hyperbolic models, we have $\mathrm{aper}_h(x_h)=\mathrm{aper}_b(x_b)$. 
Substituting the above mapping into the half-aperture definition gives

\begin{equation}
\mathrm{aper}_h(x_h)=
\sin^{-1}\!\left(
\frac{2K}{\sqrt{c}\|x_h\|}
\right).
\end{equation}

Next we derive the exterior angle used to determine the entailment relation. 
Consider three points in the Lorentz model: the origin $O$, a text embedding $x$, and an image embedding $y$. 
These three points form a hyperbolic triangle whose edges are geodesics connecting each pair of points. 
Let the Lorentzian distances be $x=d(O,y)$, $y=d(O,x)$ and $z=d(x,y)$. 
Using the hyperbolic law of cosines, the exterior angle between $x$ and $y$ with respect to the origin can be written as

\begin{equation}
\mathrm{ext}(x,y)=\pi-\angle Oxy
\end{equation}

\begin{equation}
=\pi-\cos^{-1}\!\left(
\frac{\cosh(z\sqrt{c})\cosh(y\sqrt{c})-\cosh(x\sqrt{c})}
{\sinh(z\sqrt{c})\sinh(y\sqrt{c})}
\right).
\end{equation}

Using the identity $\pi-\cos^{-1}(t)=\cos^{-1}(-t)$ and defining a shorthand function $g(t)=\cosh(\sqrt{c}\,t)$ for brevity, together with the hyperbolic trigonometric identity $\sinh(t)=\sqrt{\cosh^2(t)-1}$, the above equation can be rewritten as

\begin{equation}
\mathrm{ext}(x,y)=
\cos^{-1}\!\left(
\frac{g(x)-g(z)g(y)}
{\sqrt{g(z)^2-1}\sqrt{g(y)^2-1}}
\right).
\end{equation}

We now compute the terms $g(x)$, $g(y)$ and $g(z)$. 
Using the relation between hyperbolic distance and the Lorentzian inner product $\langle\cdot,\cdot\rangle_L$,

\begin{equation}
\cosh(\sqrt{c}\,d(a,b))=-c\langle a,b\rangle_L,
\end{equation}

we obtain

\begin{equation}
g(z)=\cosh(\sqrt{c}\,d(x,y))=-c\langle x,y\rangle_L,
\end{equation}

and similarly

\begin{equation}
g(x)=-c\langle O,y\rangle_L,
\qquad
g(y)=-c\langle O,x\rangle_L.
\end{equation}

Substituting these expressions into the exterior angle formulation yields a representation of $\mathrm{ext}(x,y)$ purely in terms of Lorentz coordinates. 
Finally, the entailment relation is determined by comparing the exterior angle with the cone aperture centered at $x$, and the entailment constraint is satisfied when

\begin{equation}
\mathrm{ext}(x,y)\leq \mathrm{aper}_h(x).
\end{equation}

\section{Additional Analysis}
\subsection{More Ablations}
\begin{table}[h]
\caption{Effect of the InfoNCE temperature $\tau$ on HSG. All experiments use the same hyper-parameters as the base HSG model, with only the temperature parameter varied.}
\label{tab:temperature}
\centering
\small
\begin{tabular}{c|cccc}
\hline
Temperature $\tau$ & PP IoU & PO IoU & Graph IoU & Recall \\
\hline
0.50 & 0.1550 & 0.4108 & 0.1574 & 0.9806 \\
0.20 & 0.3082 & 0.4478 & 0.3116 & 0.9839 \\
0.12 & 0.3305 & 0.4542 & 0.3340 & 0.9840 \\
0.10 (default) & \textbf{0.3317} & \textbf{0.4552} & \textbf{0.3351} & 0.9839 \\
0.08 & 0.3311 & 0.4501 & 0.3346 & \textbf{0.9842} \\
0.05 & 0.3309 & 0.4497 & 0.3343 & 0.9841 \\
0.01 & 0.3306 & 0.4194 & 0.3327 & 0.9841 \\
\hline
\end{tabular}
\end{table}

\noindent\textbf{Effect of temperature $\tau$.} Table~\ref{tab:temperature} studies the effect of the InfoNCE temperature $\tau$ on HSG while keeping all other hyper-parameters unchanged. 
Performance improves significantly when decreasing $\tau$ from 0.5 to 0.2, indicating that a smaller temperature helps produce more discriminative embeddings. 
The best performance is achieved around $\tau=0.1$, which yields the highest PP IoU, PO IoU, and Graph IoU. 
When $\tau$ becomes too small (e.g., 0.01), the performance slightly degrades, suggesting that overly sharp contrastive distributions may harm representation quality. 
Overall, the results demonstrate that HSG is relatively robust to temperature choices within the range $[0.05, 0.12]$, while $\tau=0.1$ provides the best overall performance. 

Empirically, smaller temperatures tend to produce stronger gradients in contrastive learning, which may benefit from a smaller learning rate to maintain stable optimization. 
This observation is consistent with our results, where moderate temperatures around $\tau=0.1$ achieve the best performance under the base learning rate.

\begin{table}[h]
\centering
\caption{Effect of the embedding dimension on HSG with curvature fixed to $c=80$. 
All other hyper-parameters remain the same as the base HSG model.}
\label{tab:embedding_dim}
\small
\begin{tabular}{c|cccc}
\hline
Embedding Dim & PP IoU & PO IoU & Graph IoU & Recall \\
\hline
64  & 0.3011 & 0.3846 & 0.3032 & 0.9733 \\
128 & 0.2897 & 0.4032 & 0.2926 & 0.9667 \\
256 & 0.3102 & 0.4054 & 0.3128 & 0.9680 \\
512 & 0.3198 & 0.4317 & 0.3229 & 0.9671 \\
1024 (default) & \textbf{0.3317} & \textbf{0.4552} & \textbf{0.3351} & \textbf{0.9839} \\
2048 & 0.2953 & 0.4287 & 0.2986 & 0.9692 \\
\hline
\end{tabular}
\end{table}

\noindent\textbf{Effect of embedding dimensions:} Table~\ref{tab:embedding_dim} studies the effect of the embedding dimension while keeping the curvature fixed to $c=80$ and all other hyper-parameters identical to the base HSG configuration. 
We observe that increasing the embedding dimension from 128 to 1024 consistently improves performance across PP IoU, PO IoU, and Graph IoU, indicating that larger embedding spaces better capture the hierarchical structure of scene graphs. 
The best performance is achieved at dimension 1024, which provides the highest scores across all metrics. 
However, further increasing the dimension to 2048 leads to noticeable performance degradation, suggesting that overly large embedding spaces may introduce redundancy and make optimization more difficult. 
These results indicate that a moderate embedding dimension offers a good balance between representation capacity and training stability for HSG.

\section{More Visualisations}
\subsection{Projecting onto the Poincar\'e ball}
\begin{figure}[h]
\centering
\begin{subfigure}[t]{0.45\linewidth}
\centering
\includegraphics[width=\linewidth]{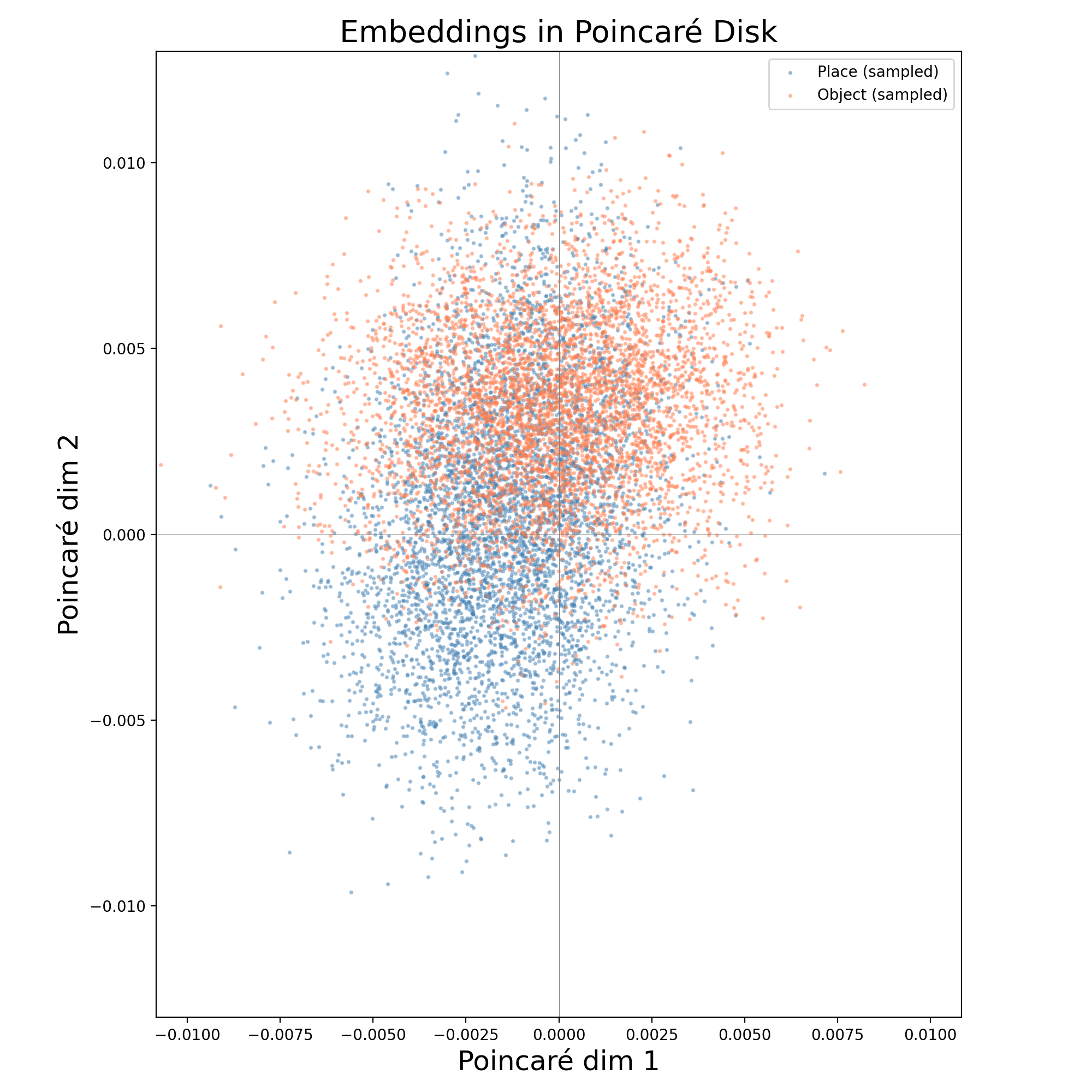}
\caption{Stage 1 (Untrained)}
\end{subfigure}
\hfill
\begin{subfigure}[t]{0.45\linewidth}
\centering
\includegraphics[width=\linewidth]{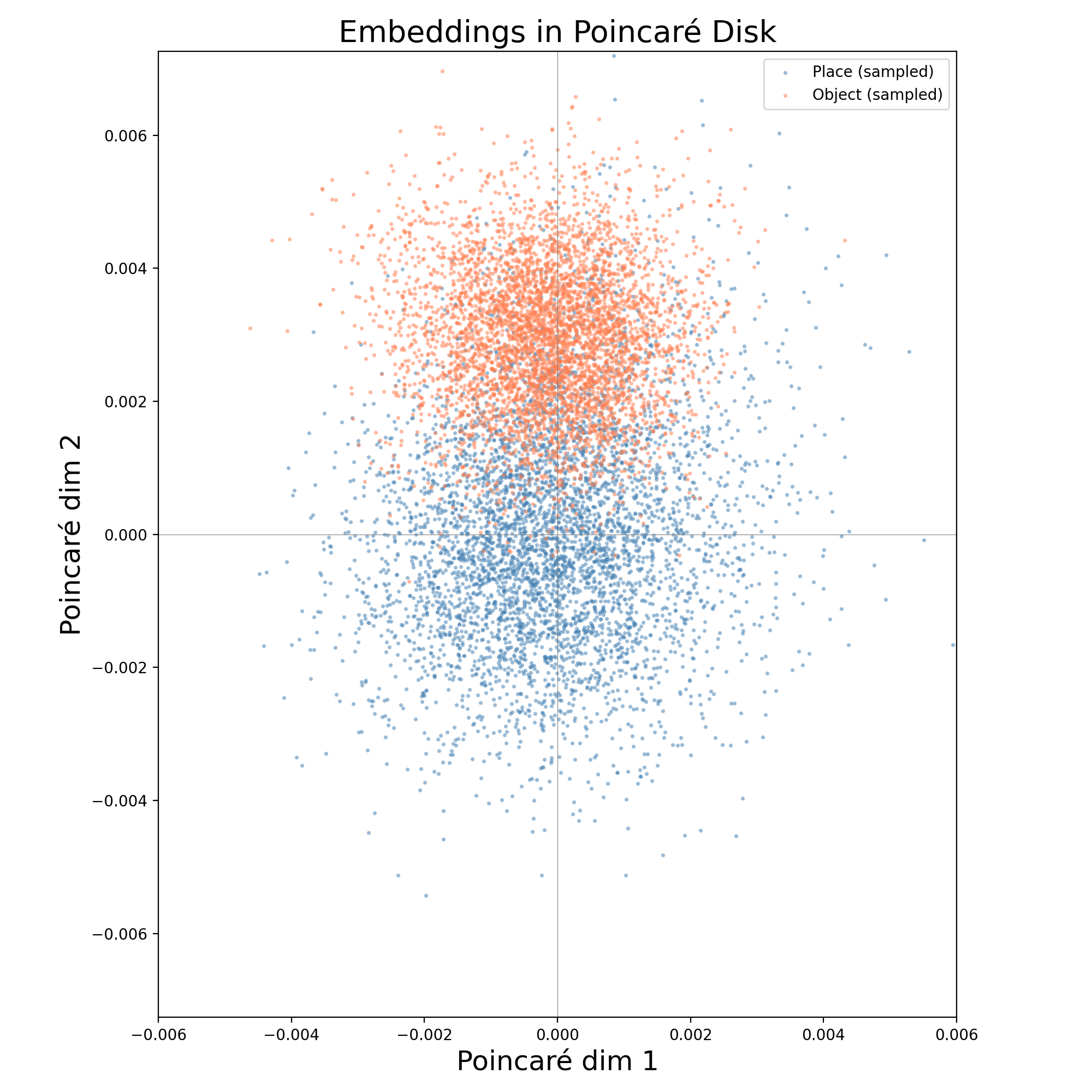}
\caption{Stage 2}
\end{subfigure}

\vspace{0.3em}

\begin{subfigure}[t]{0.32\linewidth}
\centering
\includegraphics[width=\linewidth]{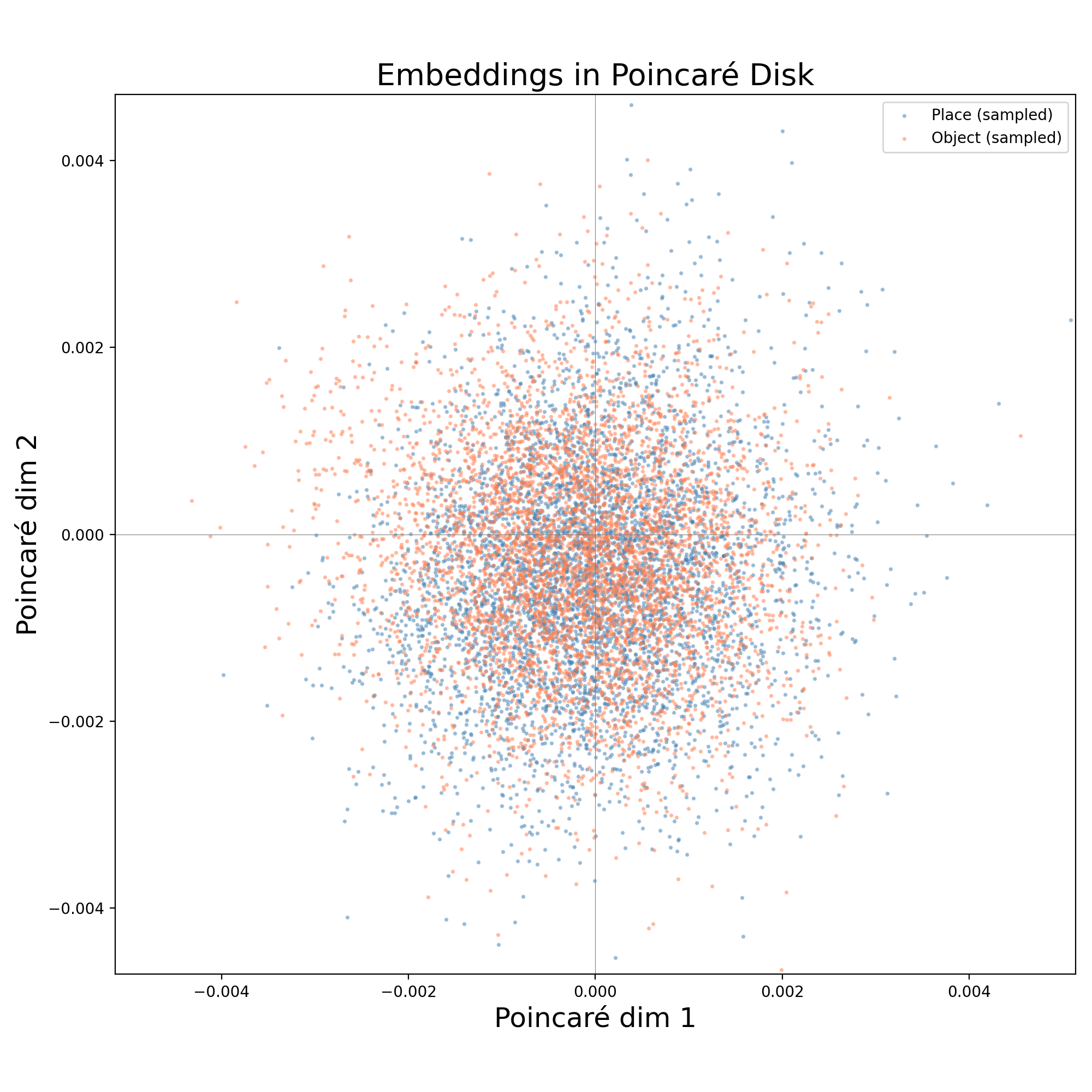}
\caption{Stage 3}
\end{subfigure}
\hfill
\begin{subfigure}[t]{0.32\linewidth}
\centering
\includegraphics[width=\linewidth]{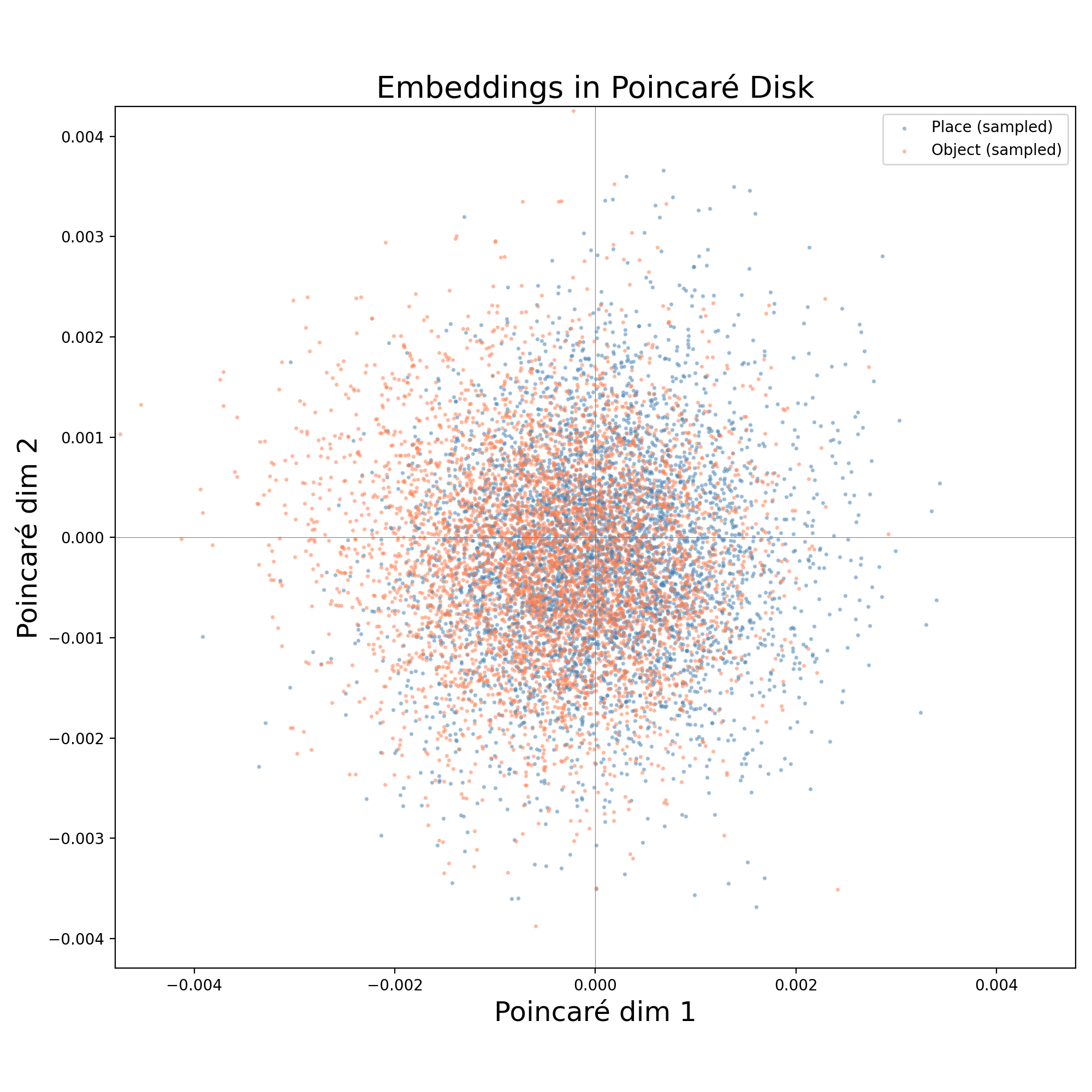}
\caption{Stage 4}
\end{subfigure}
\hfill
\begin{subfigure}[t]{0.32\linewidth}
\centering
\includegraphics[width=\linewidth]{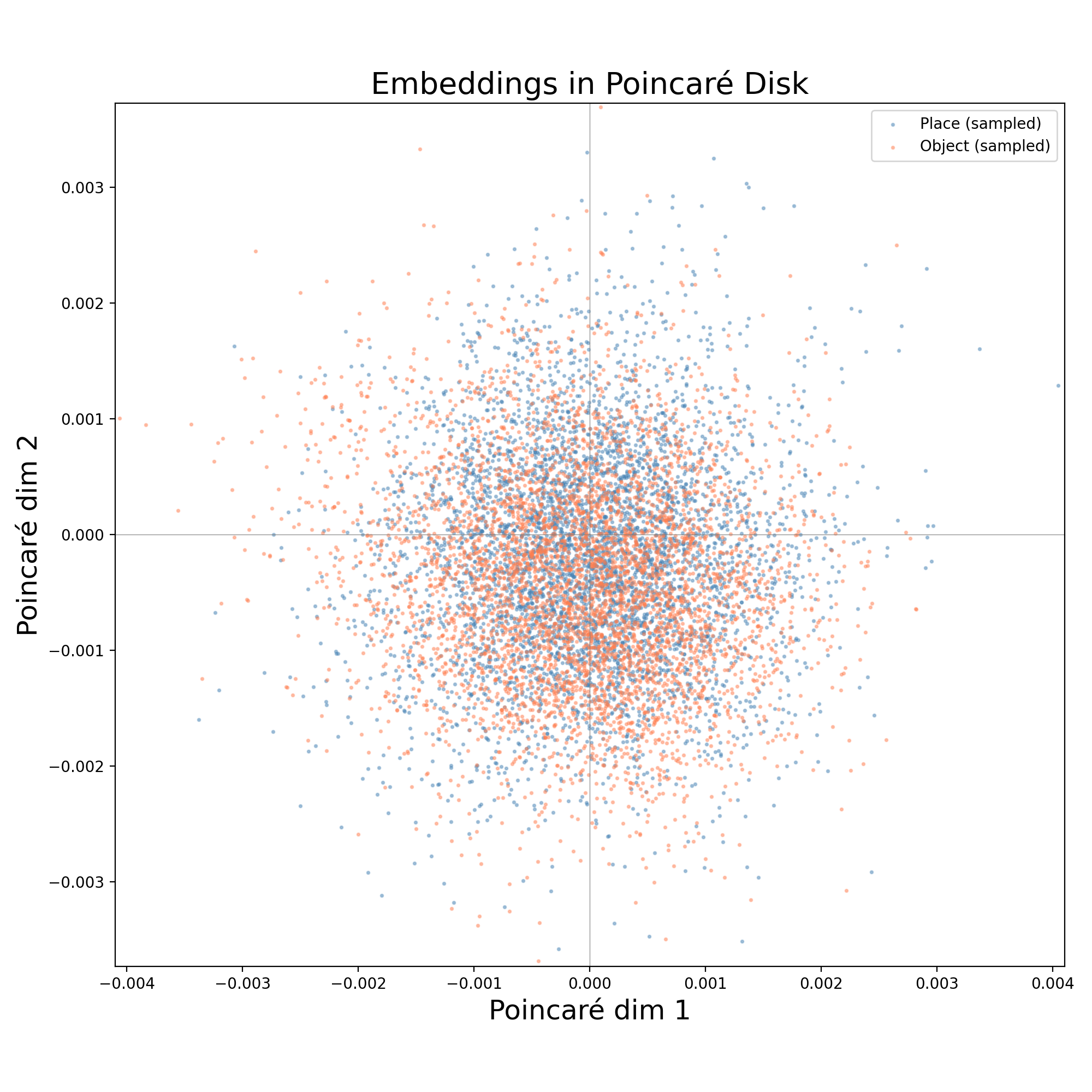}
\caption{Stage 5 (Trained)}
\end{subfigure}

\caption{Visualization of place and object embeddings projected to the Poincaré disk during training for HSG. Each stage corresponds to an evaluation performed at equal training intervals. Blue and red points correspond to place and object embeddings, respectively.}
\label{poincare_disk}
\end{figure}

Figure~\ref{poincare_disk} shows the evolution of embeddings in the Poincaré disk during training. 
At early stages, place and object embeddings are largely mixed near the center of the disk. 
As training progresses, place embeddings gradually move toward the origin while object embeddings remain relatively farther away. 
This separation leads to a hierarchical organization where place embeddings occupy more central regions of the hyperbolic space and tend to entail object embeddings. 
This behavior is consistent with the embedding distance histograms presented earlier, which show that place embeddings have smaller radial distances than object embeddings.

\subsection{Distribution of embedding distances from root}
\begin{figure}[htbp]
    \centering
    \begin{minipage}{0.48\textwidth}
        \centering
        \includegraphics[width=\linewidth]{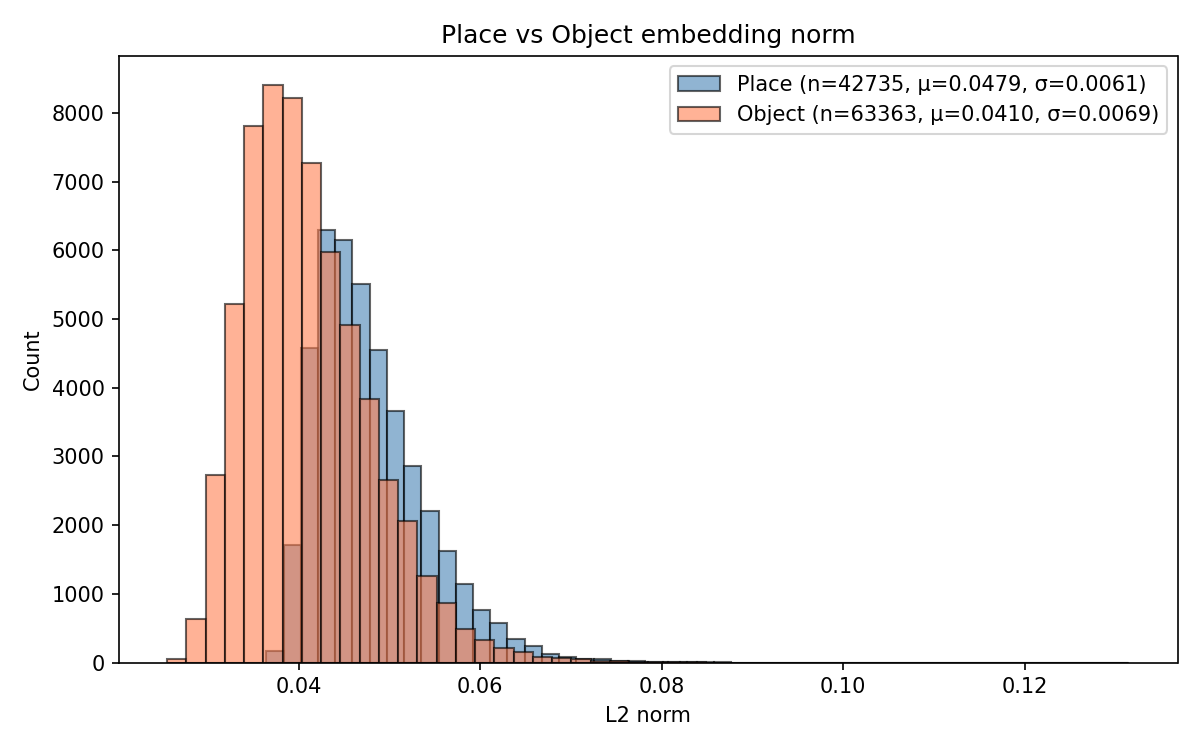}
        \caption*{(a) ConvNeXt-base}
    \end{minipage}
    \hfill
    \begin{minipage}{0.48\textwidth}
        \centering
        \includegraphics[width=\linewidth]{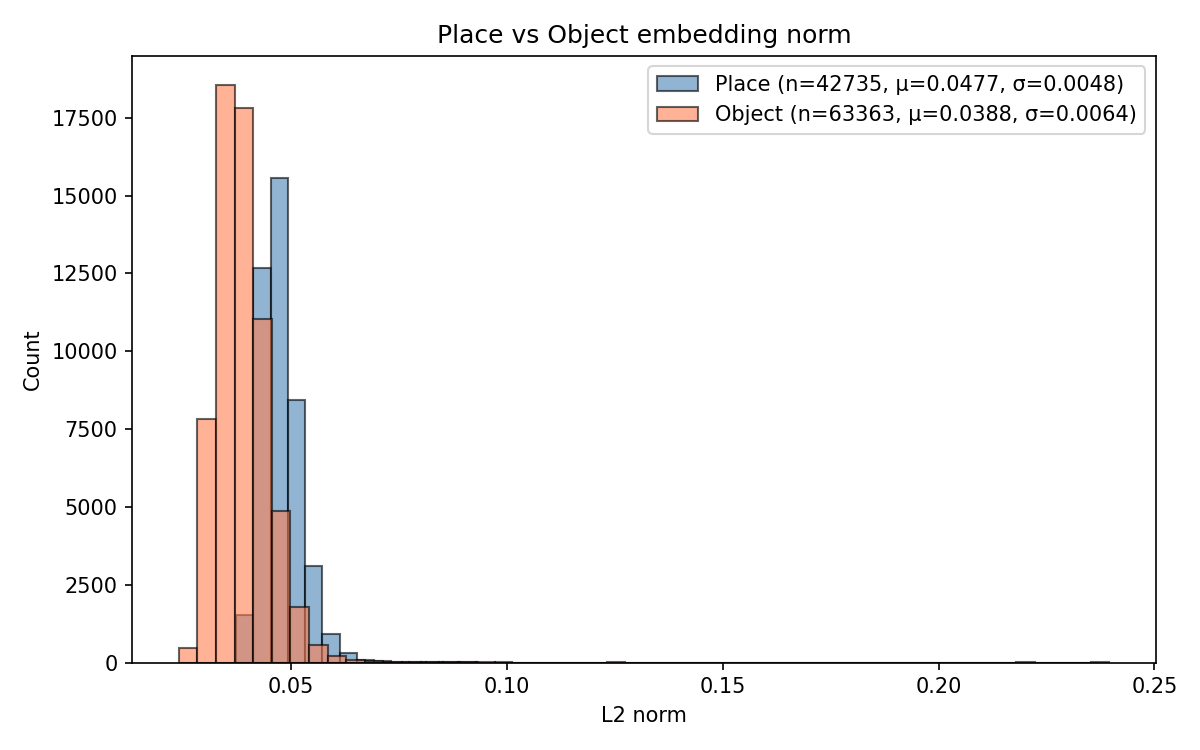}
        \caption*{(b) ViT-base}
    \end{minipage}

    \vspace{0.5em}
    \begin{minipage}{0.48\textwidth}
        \centering
        \includegraphics[width=\linewidth]{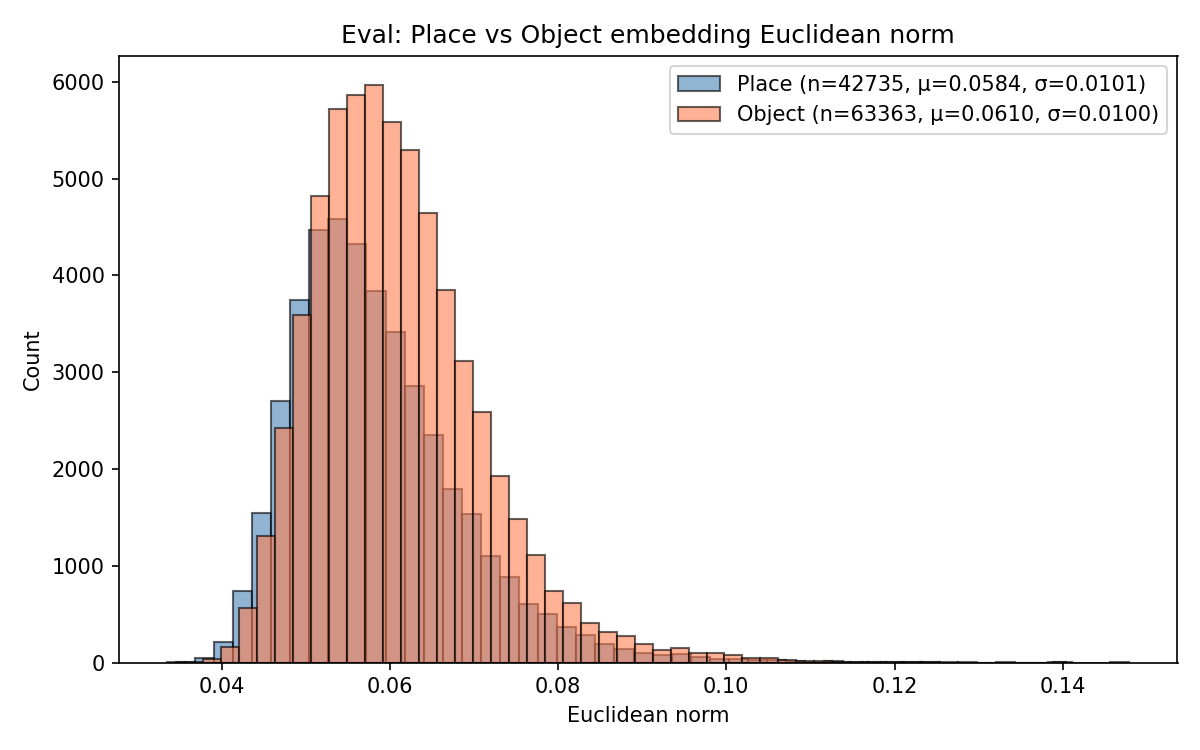}
        \caption*{(c) DINOv2-base}
    \end{minipage}

    \caption{Distribution of embedding distances from [ROOT]: We embed training images for HSG using different encoder backbones: DINOv2(default), ConvNeXt-base and ViT-base, respectively.}
    \label{fig:embedding_distance}
\end{figure}
Embedding distances from \textsc{[Root]}. To qualitatively evaluate the 
hierarchical structure learned by different encoder backbones, we visualize 
the distribution of embedding distances to the root node for HSG models 
instantiated with DINOv2-base, ConvNeXt-base, and ViT-base 
(Fig.\ref{fig:embedding_distance}). In hyperbolic space, proximity to the root reflects semantic 
generality, with more abstract concepts expected to reside closer to the 
origin. Under our formulation, \textit{place} is more 
general than \textit{object} and should therefore be embedded closer to 
the root.

As shown in Figure \ref{fig:embedding_distance}(c), the DINOv2-base encoder produces partially 
overlapping but distinguishable distributions, with place embeddings 
($\mu{=}0.0584$) exhibiting slightly smaller root distances than object 
embeddings ($\mu{=}0.0610$), suggesting a mild but consistent hierarchical ordering. ConvNeXt-base (Figure \ref{fig:embedding_distance}(a)) shows a more pronounced separation, though the ordering is inverted relative to 
the expected structure, indicating limited alignment with the intended 
hierarchy. ViT-base (Figure \ref{fig:embedding_distance}(b)) produces heavily 
overlapping distributions ($\mu{=}0.0477$ vs.\ $\mu{=}0.0388$) with 
comparable spread, reflecting weak hierarchical organization. 

\subsection{Qualitative real-world experiment}
\begin{figure*}[t]
\centering

\includegraphics[width=0.75\textwidth]{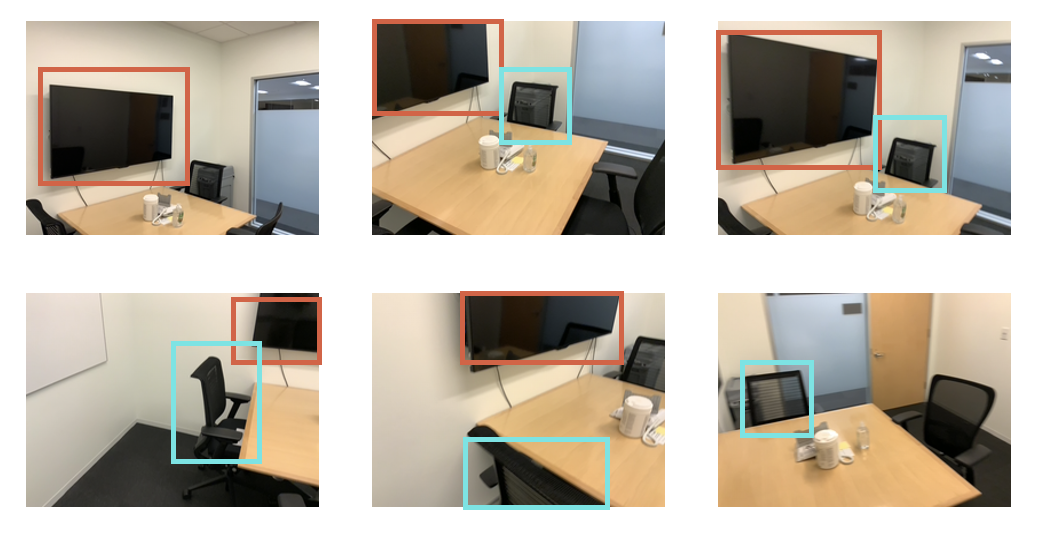}

\vspace{0.5em}

\includegraphics[width=0.83\textwidth]{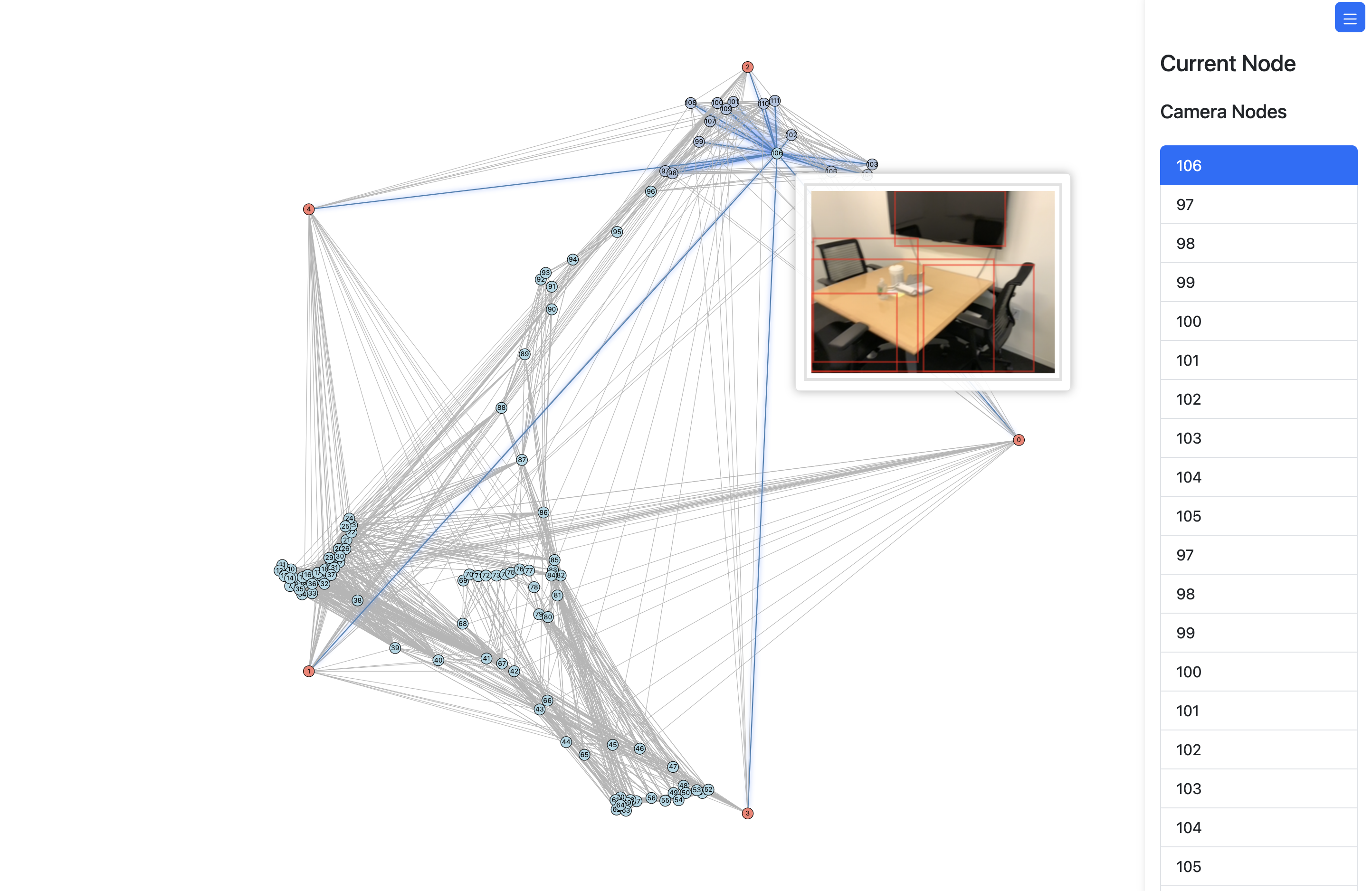}

\caption{Top: visualization of detection results. Bottom: a screenshot of the interactive graph visualization, where place nodes are shown in blue and object nodes in orange.}
\label{fig:node_visualisations}
\end{figure*}

Figure \ref{fig:node_visualisations} shows an example of a real-world experiment using the trained baseline with a pretrained
Grounding DINO detector, illustrating the connections between places and objects. The results indicate that the model produces coherent and meaningful outputs on videos beyond the training dataset.

\subsection{Other visualisations}
\begin{figure*}[t]
\centering

\includegraphics[width=0.62\textwidth]{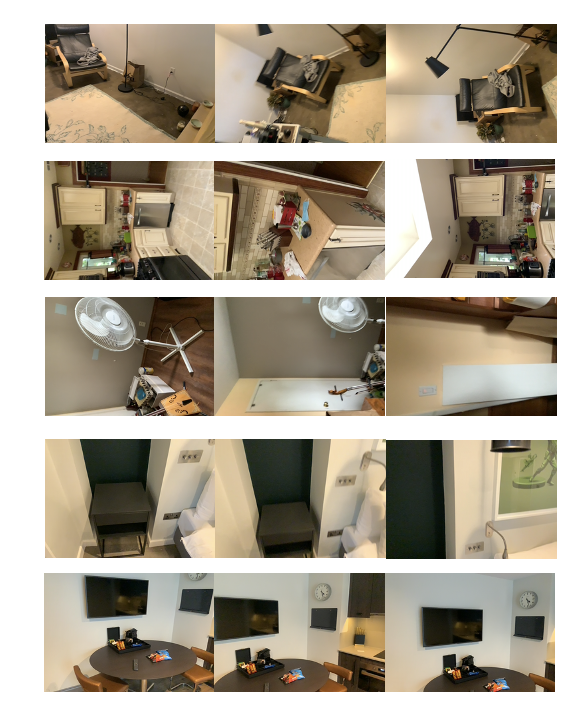}

\includegraphics[width=0.6\textwidth]{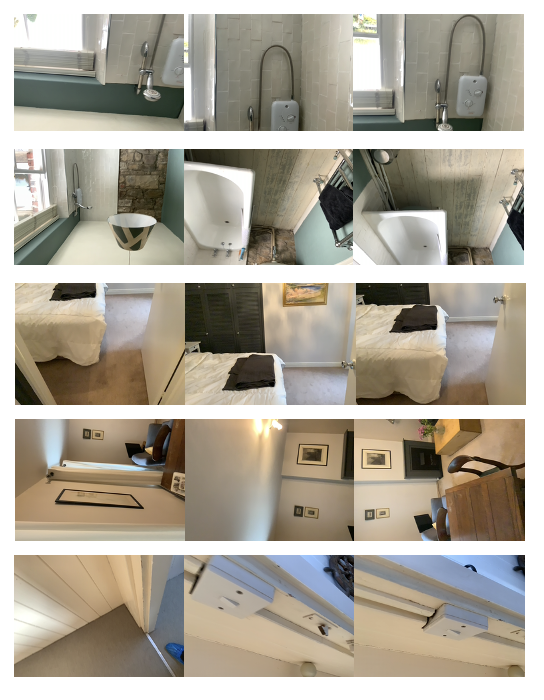}

\caption{Visualization of place nodes. Each group of three images shown side by side corresponds to nodes connected in the MSG, indicating that they are recognized as the same place.}
\label{fig:place_nodes}
\end{figure*}

\begin{figure*}[t]
\centering

\includegraphics[width=0.62\textwidth]{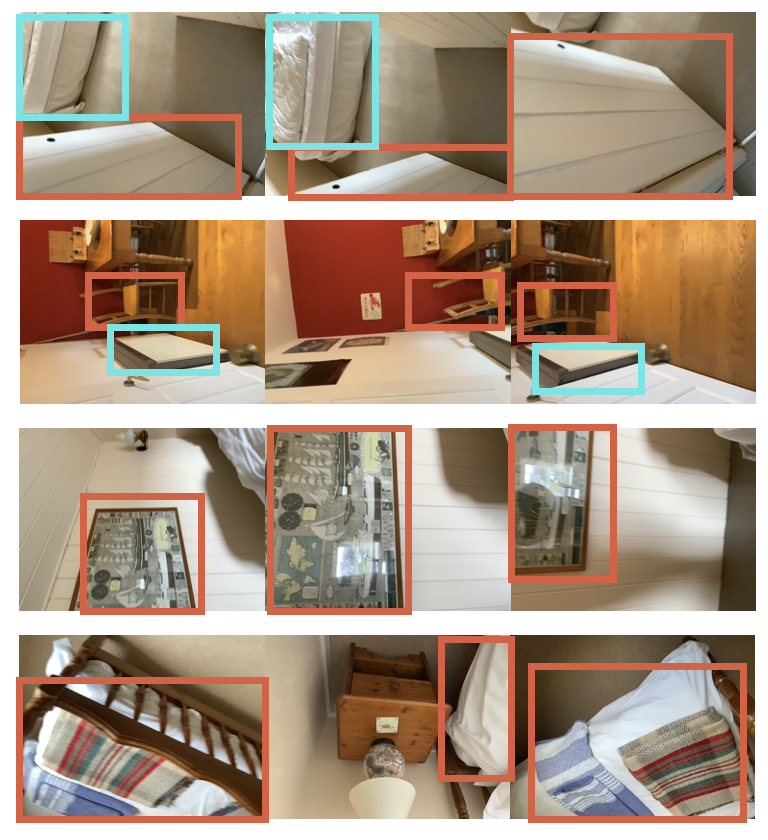}

\includegraphics[width=0.62\textwidth]{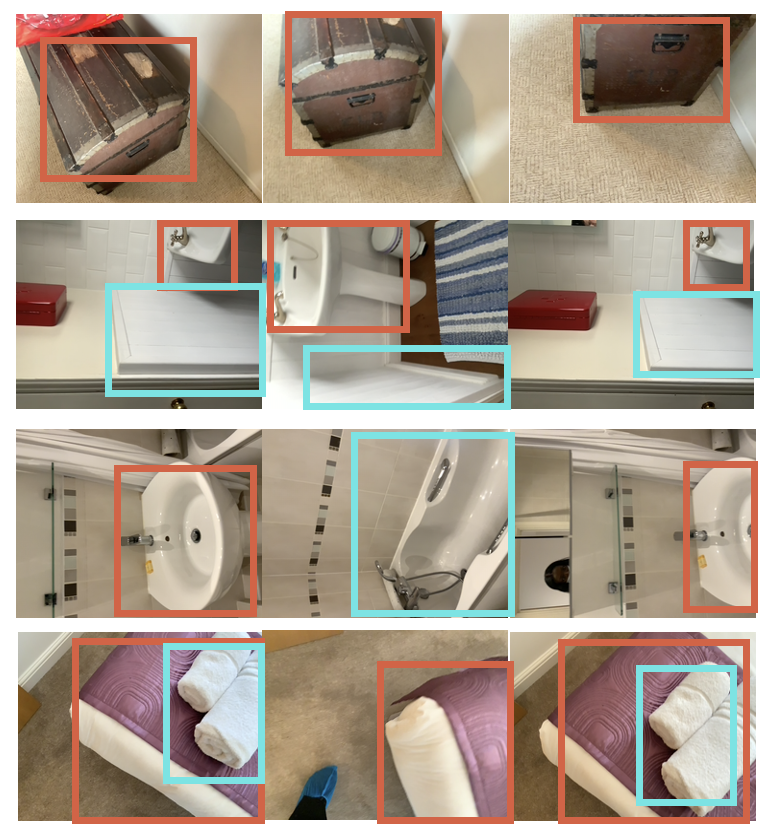}

\caption{Objects recognized across different views that correspond to the same physical object are grouped into a single object node. Within a scene, instances of the same object are highlighted with the same color. Note that some images were captured sideways; in the visualization, we retain their original orientation.}
\label{fig:place_nodes_2}
\end{figure*}

\end{document}